%% file: iclr2027_arxiv.tex
\documentclass{article}
\PassOptionsToPackage{table}{xcolor}
\usepackage{iclr2027_conference,times}
\usepackage{hyperref}

\input{math_commands.tex}

\usepackage{url}
\usepackage{tikz,lipsum}
\usepackage[most]{tcolorbox}

\tcbset{splitbox/.style={enhanced jigsaw,breakable,
  overlay first={\draw[dashed,black!25,line width=0.6pt]
    (frame.south west)--(frame.south east);},
  overlay middle={\draw[dashed,black!25,line width=0.6pt]
    (frame.north west)--(frame.north east)
    (frame.south west)--(frame.south east);},
  overlay last={\draw[dashed,black!25,line width=0.6pt]
    (frame.north west)--(frame.north east);}}}

\usepackage{amsmath,amssymb,amsfonts,mathtools,bm}
\usepackage{booktabs,tabularx,array,multirow}
\usepackage{graphicx}
\usepackage{enumitem}
\usepackage{microtype}
\usepackage{caption}
\usepackage{subcaption}
\usepackage{float}
\usepackage{placeins}
\usepackage{fvextra}
\usepackage{needspace}
\usepackage{threeparttable}
\definecolor{tabblue}{HTML}{1F77B4}

\newcolumntype{Y}{>{\raggedright\arraybackslash}X}

\title{Disentangling Self-Distillation: Measuring and Modeling Acquisition and Retention}

\author{
\begin{tabular}{l}
\bfseries
Luis Zuin \quad Alexis Huet\textsuperscript{*} \quad Dario Rossi \quad Zied Ben Houidi\textsuperscript{*} \\
\normalfont
Huawei Technologies Co., Ltd., Paris, France \\
\normalfont
\texttt{\{first(.mid).last\}@huawei.com} \\
\normalfont
\textsuperscript{*}Corresponding authors
\end{tabular}
}

\usepackage{adjustbox}

\hypersetup{
  pdftitle={Disentangling Self-Distillation: Measuring and Modeling Acquisition and Retention},
  pdfauthor={Luis Zuin, Alexis Huet, Dario Rossi, Zied Ben Houidi}
}
\iclrfinalcopy
\begin{document}

\maketitle

\begin{abstract}

Self-distillation with privileged context adapts a language model from demonstrations by letting the model, once conditioned on a reference response, teach its context-free copy token by token. 
Our taxonomy reveals existing methods differ along three entangled axes: (i)~the \textit{rollout source} (student or teacher), (ii)~the \textit{teacher coupling} 
(frozen, or an exponential moving average of the student at some coupling rate)
and (iii)~the \textit{token-level divergence} (here, reverse or forward KL), yet these axes are usually studied in fixed combinations and have led to conflicting conclusions. 
We formalize a unifying framework that covers both self-distillation and classic supervised fine-tuning:
we train every combination of the three axes, on Qwen2.5-7B and Ministral-3-3B across ordinary and contradictory tasks, totaling 1,200 adaptation runs, to systematically investigate the impact of the above axes. We propose a controlled model of the same objective to interpret the resulting acquisition--retention trade-offs.
We find that (i) the rollout source matters mostly where the task contradicts the pretrained behavior: there teacher rollouts raise acquisition (the accuracy gained on the new task) well above what student rollouts achieve, with almost no change in retention (the change in accuracy on general benchmarks); (ii) the teacher coupling changes acquisition most, on every task: acquisition rises with the coupling rate, then falls past a task-specific rate; 
(iii) switching the KL direction costs retention in one model but not the other, so which axis to tune first depends on the model. The controlled model reproduces the three trends.

\end{abstract}

\section{Introduction}\label{sec:introduction}

Adapting a deployed language model to a new task from demonstrations has long relied on supervised fine-tuning (SFT), which is known to erode prior capabilities as it learns~\citep{luo2023forgetting,yang2024sdft}. Reinforcement learning forgets less, an effect attributed to training on the model's own samples rather than on a fixed corpus~\citep{chen2025retaining,shenfeld2025razor}, but it needs a reward and delivers a sparse signal. Self-distillation with privileged context has emerged as a middle ground. In this approach, the same model, once conditioned on a demonstration, a verified answer, a document or environment feedback, becomes a dense token-level teacher for its context-free copy~\citep{snell2022context,shenfeld2026sdft,zhao2026opsd,hubotter2026sdpo,ye2026opcd,wang2026skillsd}.

This set of methods now appears under several names~\citep{song2026opdsurvey} and, in the context of this paper, we refer to it as \textit{self-distillation} for simplicity. 
Our first contribution (Sec.~\ref{sec:taxonomy}) is to show that existing methods (reviewed in Tab.~\ref{tab:related}) can be written as a single objective and differ along three main axes: the (i) \emph{rollout source}, whether the responses the student is trained on are generated by the student itself or by the context-conditioned teacher (student rollouts or teacher rollouts); the (ii) \emph{teacher coupling}, whether the self-teacher is frozen at initialization or follows the student, typically through an exponential moving average (EMA) at some coupling rate; 
and the (iii) \emph{token-level divergence}, the distillation loss on those rollouts between the teacher's and the student's next-token distributions---here, reverse or forward Kullback--Leibler (KL).
One would expect each axis to set how far training drifts from the pretrained model: student rollouts, a frozen teacher and reverse KL keep it close, teacher rollouts, a moving teacher and forward KL move it away.

Outside self-distillation, GKD~\citep{agarwal2024gkd} and Decoupling KL~\citep{zhao2026decoupling} already separate the rollout source from the KL direction; with their fixed external teacher, the coupling axis does not arise. In self-distillation, joint analysis of these axes remains limited: no study yet varies the three axes together, and most fix each of them, often implicitly (Sec.~\ref{sec:background}).
Worse yet, conclusions found in the literature often conflict (Sec.~\ref{sec:background}).
We argue that, in search of a single best method, the literature lacks an overall view of the three axes. We therefore set out to build a taxonomy of methods within the self-distillation family and provide a unifying framework (Sec.\ref{sec:taxonomy}). We then lay out a systematic evaluation plan  (Sec.\ref{sec:experimental_design}) that we carry out with complementary controlled modeling (Sec.~\ref{sec:toy_model}) and experimental (Sec.~\ref{sec:llm_results}) approaches. This overall view allows us to understand when each setting favors retention or acquisition, providing conclusions that are less sharp, but more complete, than any single study. Our contributions are the taxonomy with its single objective (Sec.~\ref{sec:taxonomy}), the systematic evaluation on LLMs (Secs.~\ref{sec:experimental_design} and~\ref{sec:llm_results}), and the controlled model (Sec.~\ref{sec:toy_model}).

We train every combination of the three axes (Sec.~\ref{sec:experimental_design}): two rollout sources, a frozen teacher or four coupling rates, two KL directions. We do so on two instruction-tuned models and five tasks, with SFT on the same demonstrations as baseline, for over 1,200 runs\footnote{Code, configurations and all run results: \href{https://github.com/lhzuin/disentangling_self_distillation}{GitHub repository}.}. We measure \emph{acquisition}, the accuracy gained on the new task, and \emph{retention}, the change in accuracy on a suite of general benchmarks.
Two of the five tasks are \emph{contradictory}: their rule opposes one the pretrained model already follows, such as arithmetic in base 9 instead of base 10. The other three are \emph{ordinary}: they refine what the model already does. A small controlled model of the same objective (Sec.~\ref{sec:toy_model}) is swept along the same axes, to see which of the LLM patterns a minimal mechanism reproduces.

The results (Sec.~\ref{sec:llm_results}) show that each of the three axes matters under a different condition. 
The teacher coupling matters on every task. Acquisition rises with the coupling rate, then falls past a rate that depends on the task and the model. 
A frozen teacher and a moving one, reported as better~\citep{shenfeld2026sdft,hubotter2026sdpo} and as worse~\citep{kim2026degrade} in the literature, can thus each be the better one, depending on where a study's rate falls.
The rollout source, by contrast, matters mostly on the contradictory tasks: there, teacher rollouts raise acquisition much more than on the ordinary tasks. 
The KL direction matters least: forward KL learns slightly more, but its advantage and retention cost are model-dependent. 
The controlled model reproduces these trends, and shows where the advantage of teacher rollouts stops.
Our baseline, SFT, still reaches the highest acquisition on most tasks, but pays for it in retention: most self-distillation settings retain more at the acquisition they reach, so they win on the trade-off, but not on acquisition.

\input{related_work/related_work}

\input{taxonomy/taxonomy}

\input{experimental_design/experimental_design}

\input{toy_model/toy_model}

\input{llm_results/experimental_results}

\input{conclusion/conclusion}

\clearpage

\subsection*{AI use statement}
In this work, we used generative AI tools to generate synthetic data sets, help develop theoretical models or conceptual frameworks, 
refine hypotheses, design or provide feedback on research methodology or experiments, implement methods, assist with translation, clean and reformat dataset, support qualitative and thematic data analysis, and interpret results. 
The rest of the required AI disclosure tasks (formulate mathematical claims, provide critical ingredients for proving mathematical claims, assist in the writing of proofs) are not applicable to this work. Additionally, we used generative AI tools to create or edit software code, summarize or analyze existing literature, identify relevant literature, and propose a title or keywords for a research paper. We have reviewed all AI-assisted work. We checked the generated literature summaries against the original papers for building the related work taxonomy. Generative AI tools were used for generating the synthetic datasets mostly to assess the existing knowledge of the models and to check for unnatural constructions in the produced datasets. We take responsibility for the final content of this work, including text, claims or artifacts produced with the aid of generative AI.

\subsection*{Reproducibility statement}
The code, produced datasets, and results are shared through the
\href{https://github.com/lhzuin/disentangling_self_distillation}{GitHub repository}.
This includes the new datasets produced in the work (math-contradiction, spatial, spatial-contradiction), for which the size, templates, examples and generation process are available in Apps.~\ref{app:datasets_contexts} and~\ref{app:llm:templates}. The code and results regarding the small controlled model are available, and a detailed description is provided in App.~\ref{toy_experimental_settings}. 
The code and results for the LLM experiments are available in the GitHub repository, and the experimental setup is described in Secs.~\ref{sec:taxonomy} and~\ref{sec:experimental_design}.

\bibliography{iclr2027_conference}
\bibliographystyle{iclr2027_conference}

\input{appendix/appendix}

\input{discussion/discussion}
\end{document}

%% file: math_commands.tex
\usepackage{amsmath,amsfonts,bm}

\def\eqref#1{equation~\ref{#1}}

\def\1{\bm{1}}

\DeclareMathAlphabet{\mathsfit}{\encodingdefault}{\sfdefault}{m}{sl}
\SetMathAlphabet{\mathsfit}{bold}{\encodingdefault}{\sfdefault}{bx}{n}

%% file: related_work/related_work.tex
\section{Related Work}\label{sec:background}

Knowledge distillation trains a student to imitate a teacher's outputs. On-policy distillation obtains this supervision along student-generated trajectories, reducing the mismatch between training and inference~\citep{agarwal2024gkd}. Self-distillation uses the model itself, or a copy of it, as teacher. In contextualized self-distillation, privileged context elicits a teaching signal through in-context learning; the student learns to reproduce the resulting behavior without that context~\citep{snell2022context}. This yields a family of methods conditioned on demonstrations or solutions~\citep{yang2024sdft,shenfeld2026sdft,zhao2026opsd}, feedback~\citep{hubotter2026sdpo}, trajectory-derived skills~\citep{wang2026skillsd}, or documents~\citep{padmanabhan2026disc,stein2026gates} that we summarize in Tab.~\ref{tab:related} (organized according to a taxonomy we propose and discuss in Sec.~\ref{sec:taxonomy}).

\textbf{Inconsistencies in the state of the art.}
We first summarize inconsistencies across the literature, axis by axis.
On (i) rollout source, teacher rollouts collapse mid-training in one study~\citep{wang2026skillsd} but are identified as the primary driver of transfer in another~\citep{stein2026gates}; student rollouts are found to protect retention inconsistently in~\citep{liu2026mixsd} yet are credited with it by~\citep{chen2025retaining}. On (ii) teacher coupling, a moving teacher is reported to outperform a frozen one~\citep{shenfeld2026sdft}, to outperform it at a slow rate but diverge when synchronized with the student~\citep{hubotter2026sdpo}, to degrade performance relative to a frozen one~\citep{kim2026degrade}, and to accumulate the student's drift without bound~\citep{guo2026teachermovetemporalcoupling}. 
On (iii) the token-level divergence, the same retention is explained instead by reverse KL~\citep{balasubramanian2026forgetting}, yet
self-distillation studies testing both directions favor forward KL~\citep{shenfeld2026sdft,zhao2026opsd}.

\textbf{Lack of an overall view.}
Tab.~\ref{tab:related} shows why these results cannot be reconciled: a study typically
varies just a single axis, fixing others with arbitrary choice and evaluating on
its own task.  Few studies jointly examine more than one axis: namely, rollout source with KL
direction under an external teacher~\citep{agarwal2024gkd,zhao2026decoupling,zhang2026prefixopd} (hence outside the self-distillation area), or rollout source with teacher coupling in Skill-SD~\citep{wang2026skillsd}. 
Besides, most studies focus on acquisition gains, whereas knowledge retention remains less studied, hiding potential serious side effects.
Interactions between axes, their dependence on the task, and their side effects therefore remain largely unknown and, in our opinion, deserve systematic evaluation.

\begin{table}[t]
\centering
\caption{\textbf{Taxonomy of related work}. For each study, the table reports individual settings used:  along a single axis, entries separated by a slash symbol (/) are \emph{compared} within that study; axes marked in \textbf{bold} are  \emph{jointly analyzed} in one reported experiment.
Retention denotes whether possible degradation is evaluated on a broad, task-unrelated capability suite.
Per-row details are deferred to App.~\ref{app:detailed_related_work}.}
\label{tab:related}
\vspace{-0.2cm}
\begingroup
\footnotesize
\setlength{\tabcolsep}{3pt}
\renewcommand{\arraystretch}{1.12}
\setlength{\aboverulesep}{0.3ex}
\setlength{\belowrulesep}{0.45ex}

\newcommand{\grp}[1]{%
  \hiderowcolors
  \multicolumn{5}{@{}l}{\emph{#1}}\\[1pt]
  \showrowcolors
}

\rowcolors{2}{gray!10}{white}

\begin{tabular}{@{}
  >{\raggedright\arraybackslash}p{85pt}
  >{\raggedright\arraybackslash}p{68pt}
  >{\raggedright\arraybackslash}p{95pt}
  >{\raggedright\arraybackslash}p{90pt}
  >{\centering\arraybackslash}p{35pt}@{}}

\toprule
\hiderowcolors
Study & Rollout source & Teacher coupling & Token-level loss & Retention\\
\midrule
\showrowcolors

\grp{Distillation from a fixed external teacher}
GKD~\citeyearpar{agarwal2024gkd}
  & \textbf{student\,/\,data} & external, fixed
  & \textbf{forward\,/\,reverse\,/\,JSD} & \\
Decoupling KL~\citeyearpar{zhao2026decoupling}
  & \textbf{student\,/\,teacher} & external, fixed
  & \textbf{forward\,/\,reverse} & $\checkmark$\\
Prefix OPD~\citeyearpar{zhang2026prefixopd}
  & \textbf{student\,/\,data} & external, fixed
  & \textbf{forward\,/\,reverse} & \\

\hiderowcolors
\midrule
\showrowcolors

\grp{Self-teacher without privileged context}
\citet{guo2026teachermovetemporalcoupling}
  & student & frozen\,/\,EMA\,/\,refresh & JSD & \\

\hiderowcolors
\midrule
\showrowcolors

\grp{Self-teacher conditioned on privileged context}
\citet{yang2024sdft}
  & teacher & frozen & hard targets & $\checkmark$\\
SDFT~\citeyearpar{shenfeld2026sdft}
  & student$^{\ddagger}$ & frozen\,/\,EMA\,/\,sync.\ & forward & $\checkmark$\\
OPSD~\citeyearpar{zhao2026opsd}
  & student & frozen & forward\,/\,reverse\,/\,JSD & \\
SDPO~\citeyearpar{hubotter2026sdpo}
  & student & frozen\,/\,EMA\,/\,sync.\,/\,TR & JSD & $\checkmark$\\
OPCD$^{\dagger}$~\citeyearpar{ye2026opcd}
  & student$^{\ddagger}$ & frozen\,/\,sync.\ & reverse$^{\ddagger}$ & $\checkmark$\\
Skill-SD~\citeyearpar{wang2026skillsd}
  & \textbf{student\,/\,teacher} & \textbf{frozen\,/\,sync.} & reverse & \\
\citet{kim2026degrade}
  & student$^{\ddagger}$ & frozen\,/\,EMA & JSD; hard targets & \\
DiSC~\citeyearpar{padmanabhan2026disc}
  & data & frozen & forward & $\checkmark$\\
MixSD~\citeyearpar{liu2026mixsd}
  & teacher $+$ base & frozen & hard targets, mixed & $\checkmark$\\
GATES~\citeyearpar{stein2026gates}
  & teacher $+$ student & sync.\ & hard targets $+$ reverse & \\

\hiderowcolors
\midrule
\showrowcolors

\textbf{Ours}
  & \textbf{student\,/\,teacher} & \textbf{frozen\,/\,EMA, 4 rates}
  & \textbf{forward\,/\,reverse} & $\mathbf{\checkmark}$\\

\hiderowcolors
\bottomrule

\end{tabular}

\vspace{3pt}
{\scriptsize\raggedright
\emph{Rollout source}: \emph{data} = prefixes read from a fixed corpus (not generated); \emph{base} = same model without context; $+$ denotes sources used together.
\emph{Teacher coupling}: \emph{sync.}\ = teacher set is the updated student; \emph{TR} = interpolation between current and initial teacher; \emph{refresh} = periodic hard copy.
\emph{Token-level loss}: \emph{hard targets} = one-hot targets on sampled tokens, for which no direction is defined;  \textit{JSD} = Jensen Shannon Divergence;
$^{\dagger}$OPCD also reports external teachers.
$^{\ddagger}$varied only against a baseline that changes other method choices at the same time, so the axis is not isolated.
\par}
\endgroup
\vspace{-0.25cm}
\end{table}

%% file: taxonomy/taxonomy.tex
\section{Taxonomy and unifying framework}\label{sec:taxonomy}

Our first contribution is to introduce a unifying framework for organizing the methods reviewed above. In particular, we formulate a common objective for self-distillation with privileged context and classic SFT, which separates the taxonomical axes of (i) rollout source, (ii) teacher coupling and (iii) token-level divergence.

\textbf{A unifying framework.} Let us introduce the notation. A training example is an \textit{input} $x$ with a \textit{reference response} $y^{\mathrm{ref}}$, and a \textit{privileged context} $c$ that the teacher sees, but the student does not. In this paper, $c$ is the reference response itself, placed in a fixed template (Sec.~\ref{sec:experimental_design}), so that self-distillation and SFT draw on the same information and differ only in how it reaches the student.
We start from a pretrained policy $\pi_{\theta_0}$ and use it in two roles: as a \emph{student}, with parameters $\theta_i$ at optimization step $i$, trained and later evaluated on $x$ alone; and as a \emph{teacher}, with parameters $\theta_i^\star$, which additionally sees $c$ and which corrects the student token by token.
Both start from the same parameters ($\theta_0^\star=\theta_0$), so the teacher's initial advantage comes from $c$, not from a stronger model.

The training procedure samples a response $\bar y$ from a rollout distribution $\mu_i(\cdot\mid x)$; at each position $t$, teacher $q$ and student $p$ read the same prefix $\bar y_{<t}$, the tokens already produced:
\begin{equation}
q_{i,t}=\pi_{\theta_i^\star}(\cdot\mid x,c,\bar y_{<t}),\qquad
p_{i,t}=\pi_{\theta_i}(\cdot\mid x,\bar y_{<t}).
\label{eq:distributions}
\end{equation}

The objective averages a  divergence $d(q_{i,t},p_{i,t})$ over training examples $x\mkern-5.7mu\sim\mkern-5.7mu\mathcal D$ and response tokens; only the student $p_{i,t}$ receives the gradient, the teacher and the sampled response being held fixed:
\begin{equation}
\textstyle
\mathcal L_i(\theta_i)=
\mathbb E_{x\sim\mathcal D}\,
\mathbb E_{\bar y\sim\mu_i(\cdot\mid x)}
\left[
\frac{1}{|\bar y|}\sum_{t=1}^{|\bar y|} d(q_{i,t},p_{i,t})
\right].
\label{eq:objective}
\end{equation}
Three choices in Eq.~\ref{eq:objective} are left open, which map to the taxonomical axes: the rollout distribution $\mu_i$, the  divergence $d(\cdot,\cdot)$, and the teacher coupling, i.e., how $\theta_i^\star$ follows $\theta_i$. Together,  as illustrated later (Sec.~\ref{sec:experimental_design}), these decisions govern the trade-off between acquisition and retention.

\textbf{Axis 1: Rollout source.} The rollout source fixes which prefixes the student is corrected on, and  student rollouts are prevalent over teacher rollouts in the literature. With \emph{student rollouts} $\mu_i(\cdot\mid x)=\pi_{\theta_i}(\cdot\mid x)$, the student is corrected on the prefixes it produces itself.  Thus, training happens on the text the student produces at deployment, which is credited with acquisition~\cite{agarwal2024gkd} and retention~\citep{chen2025retaining}. The correction, however, is only as useful as what $c$ still adds on those prefixes~\citep{liu2026smrcsd}.
With \emph{teacher rollouts} $\mu_i(\cdot\mid x)=\pi_{\theta_i^\star}(\cdot\mid x,c)$, the teacher generates the response with $c$ in context, and the student is corrected on prefixes where the teacher is reliably informative, but that the student may never produce at deployment.

\textbf{Axis 2: Teacher coupling.}
The teacher coupling fixes how the teacher follows the student. After each student step, the teacher moves toward it by an exponential moving average with rate $\alpha\in[0,1]$:
\begin{equation}
\textstyle
\theta_{i+1}^\star=(1-\alpha)\,\theta_i^\star+\alpha\,\theta_{i+1}.
\label{eq:ema}
\end{equation}
A \emph{frozen} teacher ($\alpha=0$), prevalent in self-distillation literature, remains a fixed pretrained model with $c$ in context; 
the other extreme ($\alpha=1$) synchronizes the teacher with the updated student~\citep{shenfeld2026sdft,hubotter2026sdpo,wang2026skillsd}. In between,
an \emph{EMA} teacher ($\alpha>0$), adopted e.g. in ~\citet{shenfeld2026sdft,hubotter2026sdpo,kim2026degrade},  incorporates what the student has learned, so that its targets improve as training progresses, but also inherits the student's drift, which it then feeds back to the student~\citep{guo2026teachermovetemporalcoupling}.   Other constructions of the teacher from the current and initial models fit Eq.~\ref{eq:objective} as well, such as the trust-region teacher of SDPO~\citep{hubotter2026sdpo}, which interpolates between the current and initial teachers; our grid uses the EMA of Eq.~\ref{eq:ema}.

\textbf{Axis 3: Token-level divergence.} The divergence $d$ fixes how the disagreement at each token is weighted in Eq.~\ref{eq:objective}. Our experiments compare forward and reverse KL. With \emph{forward KL} ($d(q,p)=D_{\mathrm{KL}}(q\Vert p)$), the tokens the teacher favors are pushed up, whether or not the student already gives them mass; with \emph{reverse KL} ($d(q,p)=D_{\mathrm{KL}}(p\Vert q)$), the correction is weighted by the student's own probabilities, so that it concentrates on the teacher's modes the student already approaches~\citep{gu2024minillm}. 
Forward KL is thus mode-covering and reverse KL mode-seeking. This asymmetry matters when the teacher keeps several modes: reverse KL can then settle on the one the student already holds, minimizing impact in previous behavior,
whereas forward KL has to cover them all~\citep{balasubramanian2026forgetting,chen2025retaining}.   Other divergences fit as well, such as the Jensen--Shannon divergence~\citep{hubotter2026sdpo,kim2026degrade,guo2026teachermovetemporalcoupling,zhao2026opsd}.

\textbf{Taxonomical implications}
Even classic SFT can be expressed as a special case of Eq.~\ref{eq:objective}, as in the unified view of Decoupling KL~\citep{zhao2026decoupling}: the sampled response is the reference itself ($\bar y=y^{\mathrm{ref}}$), the teacher is the one-hot distribution on the reference token ($q_{i,t}=\delta_{y^{\mathrm{ref}}_t}$), and using forward KL the loss is the usual teacher-forced negative log-likelihood. 
As previously observed, Tab.~\ref{tab:related} places existing methods in this grid: most use student rollouts, with teacher rollouts appearing mostly in baselines, and coupling at one of the two extremes, frozen ($\alpha=0$) or synchronized ($\alpha=1$).
At a high level, student rollouts, reverse KL and a frozen teacher each hold the update closer to the original policy, while their counterparts increase adaptation capability. Further, each axis acts on a different object (which prefixes, which tokens, which teacher), whereas a task acts on everyone at once (the prefixes the student visits, its disagreement with the teacher, and what $c$ adds). 
Due to contrasting experimental results arising in the literature, we expect no universal ranking to arise.
As such, more than identifying a single best configuration, this paper aims at gathering a broader and coherent understanding of the relative importance of each axis, on ordinary and contradictory tasks, with classic SFT as a reference. 
In the rest of the paper, we do so with LLM experiments (Sec.~\ref{sec:llm_results}) compared with a controlled model of the same fine-tuning (Sec.~\ref{sec:toy_model}).

%% file: experimental_design/experimental_design.tex
\section{Design space investigation}
\label{sec:experimental_design}

Our investigation targets how rollout source, teacher coupling, and KL direction affect acquisition and retention. To do so, we cross the three axes (2 rollout sources, 5 coupling rates and 2 KL directions), into 20 self-distillation configurations, and run each over 5 tasks, 2 models, 3 learning rates and 2 seeds, for 1,200 overall training runs (cfr. Tab.~\ref{tab:grid}a), with systematic comparison to classic SFT.

\textbf{Tasks and privileged context.}
We use five tasks. Two of the ordinary tasks are inherited from the SDFT protocol~\citep{shenfeld2026sdft}: \emph{tool-alpaca}~\citep{tang2023toolalpaca} requires structured tool calls, and \emph{chemistry} (SciKnowEval Chemistry L-3,~\citealp{feng2024sciknoweval}) tests scientific reasoning through multiple-choice questions.
The three remaining tasks are new, one of which is a control task, and the remaining two are \textit{contradictory}: i.e., they turn two counterfactual evaluations of \citet{wu2024reasoning} into training tasks. In detail, each follows a rule that clashes with the one the model knows, applied consistently so that the task is learnable (Tab.~\ref{tab:contra-examples}): \emph{spatial-contradiction} composes spatial relations to recover grid coordinates, with the compass rotated; \emph{math-contradiction} rewrites DeepMind Mathematics~\citep{saxton2019mathematics} in base 9, where the model expects base 10. 
The ordinary control\footnote{We add no such control for math-contradiction, as both models already answer ordinary DeepMind Mathematics correctly, with an initial accuracy of 91.4\% for Qwen2.5-7B, 82.4\% for Ministral~3~3B.}, \emph{spatial}, differs from spatial-contradiction only in its reference solutions and answers. 
Both models start below $2\%$ accuracy on spatial-contradiction, against $20$-$26\%$ on spatial. Detailed dataset sizes and all initial accuracies are deferred to Tab.~\ref{tab:task-baselines}  in App.~\ref{app:datasets_contexts}.

For both contradictory tasks, problems and reference solutions are verified programmatically, as follows. The spatial pair is fully generated, with varied phrasing (semantics and syntax). The step-by-step solutions of math-contradiction are generated by an expert model (Qwen3-30B-A3B), filtered for correctness, and converted to base 9; an LLM judge (GPT-5-mini) then audits each full solution for base-9 consistency, and only examples judged consistent are retained. Detailed construction is available in App.~\ref{app:datasets_contexts}. 
In all tasks, the teacher receives the problem's full reference response through a fixed task-specific template; the student receives only the task prompt.
Holding the privileged context template fixed across all configurations ensures that only the three axes vary (see App.~\ref{app:llm:templates}); App.~\ref{app:llm_context_ablations} tests whether our conclusions depend on this choice by giving the teacher only the final answer, or feedback on the student's response.

\begin{table}[t]
\vspace{-0.2cm}
\centering
\footnotesize
\setlength{\tabcolsep}{4pt}
\caption{\textbf{Minimal examples of contradictory tasks.} The last two columns give the reference answer under the ordinary and the contradictory convention (see App.~\ref{example:first}--\ref{example:last} for more details)}
\label{tab:contra-examples}
\vspace{-0.2cm}
\resizebox{\textwidth}{!}{%
\begin{tabular}{@{}llm{0.38\linewidth}cc@{}}
\toprule
\bf Task & \bf Undisclosed convention & \bf Problem & \bf Ordinary & \bf Contradictory \\
\midrule
Spatial & Directions rotated $90^\circ$ clockwise & On the grid, Vito is at $(-2, 3)$. The position of Galen is one grid cell down from Vito. Determine the coordinates of Galen. & $(-2, 2)$ & $(-3, 3)$ \\
\addlinespace[2pt]
Math & Numerals in base nine & $(153 - 140) + (-3 - (-2 - (-8)))$ & $4$ & $3$ \\
\bottomrule
\end{tabular}
}
\vspace{-0.2cm}
\end{table}

\textbf{Models and training.}
We cross the three axes on two instruction-tuned models of different families and sizes, namely Qwen2.5-7B-Instruct~\citep{qwen2025qwen25} and Ministral~3~3B~Instruct~\citep{ministral3card}. 
Each task is adapted separately, always starting from the initial model. 
The 20 self-distillation configurations are compared with SFT, which minimizes cross-entropy on the same reference responses over a wider learning-rate grid\footnote{Selected in $\{$2e-6, 3.25e-6, 5e-6, 1e-5, 2e-5, 3.25e-5, 5e-5$\}$, for a total of $140$ runs.}. SFT thus sees the same reference responses that the teacher reads in context, so the two methods differ only in how the reference enters training: as the target of cross-entropy, or through the teacher's prompt.
We fine-tune all student parameters with AdamW and a cosine schedule, following the released SDFT training protocol~\citep{shenfeld2026sdft}. 
Unless stated otherwise, epochs are fixed per task: 2 for tool-alpaca and chemistry (as in the released protocol), and 4 for the three new tasks;  the learning rate is fixed per model at the largest grid value for which no individual run loses more than 5pp of retention (2e-5 for Qwen and 5e-6 for Ministral).

\textbf{Evaluation.}
We evaluate the student without privileged context, on each task's held-out validation set.
\emph{Acquisition} is the accuracy gain: the mean task accuracy over the last three saved checkpoints of each run, minus the initial accuracy.
\emph{Retention} is the mean change from the initial model over five general-capability benchmarks (HellaSwag, MMLU, TruthfulQA-MC2, WinoGrande, IFEval), evaluated with the Language Model Evaluation Harness~\citep{gao2024lmeval} at the final checkpoint; negative values indicate forgetting.
Task accuracy is evaluated at every saved checkpoint, and we average the last three to reduce checkpoint-level variance. The retention evaluation is more costly and is run at the final checkpoint only. App.~\ref{app:llm_results_same_ckpt} repeats the per-axis tests with acquisition measured at the final checkpoint.
Both metrics are computed per run, then averaged over the two seeds  (App.~\ref{app:llm_results_third_seed} adds a third seed to Figs.~\ref{fig:llm:rollout}a and~\ref{fig:llm:coupling}a).
Unless stated otherwise, LLM figures plot these two quantities in percentage points (pp), with axes labeled \emph{acquisition} and \emph{retention}. 
Significance for each pairwise comparison along an axis is assessed, per model and task, with a Wilcoxon signed-rank test over matched pairs: two runs that share learning rate and the other two axes and differ only on the tested one, averaged over seeds. Figures and tables that instead show a single configuration, at its best coupling rate $\alpha^\star$ (frozen included) or, for SFT, at its best learning rate, select it by seed-averaged validation acquisition; App.~\ref{app:llm_results_test_split} evaluates this selection on a separate test split.
Sec.~\ref{sec:toy_model} instantiates the same three axes in a finite autoregressive model; model and experimental LLM results are contrasted in Sec.~\ref{sec:llm_results}.

\newlength{\greekgap}
\setlength{\greekgap}{\dimexpr(\aboverulesep+\belowrulesep+\lightrulewidth)/2\relax}
\begin{table}[t]
\centering
\scriptsize
\setlength{\tabcolsep}{3pt}
\caption{\textbf{The experimental (a) and model (b) grids.} The three axes are
fully crossed: (a)~$2\times5\times2 = 20$ self-distillation configurations, $1200$ training
runs, epochs fixed per task; (b)~$2\times18\times2 = 72$ configurations, crossed
with the three quantities of Sec.~\ref{sec:toy_model}
(other setting in Tab.~\ref{tab:mechanistic_reference}).} 
\label{tab:grid}
\vspace{-0.2cm}
\begin{tabular}[t]{@{}c@{}}
(a) LLM experiments\\[3pt]
\begin{tabular}{@{}llc@{}}
\toprule
 & Values & \# \\
\midrule
Rollout source & student, teacher & 2 \\
Teacher coupling $\alpha$ & $0,0.02,0.05,0.10,0.25$ & 5 \\
KL direction & reverse, forward & 2 \\
\midrule
\multirow{2}{*}{Task} \emph{ordinary} & tool-alpaca, chemistry, spatial & \multirow{2}{*}{5} \\
\hphantom{Task} \emph{contradictory} & spatial-contra, math-contra & \\
\midrule
Model & Qwen2.5-7B, Ministral~3~3B & 2 \\
\midrule
Learning rate (lr) & 5e-6, 1e-5, 2e-5 & 3 \\
Seeds & & 2 \\
\bottomrule
\addlinespace[2pt]
\multicolumn{3}{r@{}}{1200} \\
\end{tabular}
\end{tabular}\hspace{40pt}
\begin{tabular}[t]{@{}c@{}}
(b) Controlled model\\[3pt]
\begin{tabular}{@{}llc@{}}
\toprule
 & Values & \# \\
\midrule
Rollout source & student, teacher & 2 \\
Teacher coupling $\alpha$ & 0, \ldots, 0.16 & 18 \\
KL direction & reverse, forward & 2 \\
\midrule
\noalign{\vskip\greekgap}
$\lambda$ initial concentration & 0.625, 0.88 \ldots 5 & 7 \\
$\kappa$ context strength & 0.3, 0.6, 0.9 & 3 \\
$\rho_\varphi$ feature overlap & 0, 0.5, 1 & 3 \\
\noalign{\vskip\greekgap}
\midrule
Learning rate (lr) & 0.001 (default) & \\
Seeds & & 96 \\
\bottomrule
\addlinespace[2pt]
\multicolumn{3}{r@{}}{435\,456} \\
\end{tabular}
\end{tabular}
\vspace{-0.6cm}
\end{table}

%% file: toy_model/toy_model.tex
\section{A Controlled Model of Self-Distillation}\label{sec:toy_model}

This section builds a small controlled model of the self-distillation of Sec.~\ref{sec:experimental_design}, in which the factors are set separately, to see which drives the effect of each axis in Sec.~\ref{sec:llm_results}. The model is the analogue of a frozen transformer whose LM head is fine-tuned: every input has a hidden state that never changes, and the head is the only trainable part. Before fine-tuning, this head produces what the model already does, which we call the old task; as there is one head for both tasks, fine-tuning on the new task moves the old behavior too. Our model adapts the continual learning models of~\citet{lee2021continual,hiratani2024disentangling}
adding (i) token-by-token generation and (ii) a teacher with privileged context.

\textbf{Model parameters.}
The controlled model depends on three parameters: (i) $\lambda$, the \emph{initial-policy concentration}, controls how confidently the initial model holds a behavior the new task opposes: large $\lambda$ is the model's counterpart of a contradictory task. (ii) $\kappa$, the \emph{context strength}, is how far the privileged context alone moves the initial model toward the new task's target behavior.
Finally (iii) $\rho_\phi$, the \emph{feature overlap}, assesses the similarity of new vs old task inputs: in the continual learning models above, it sets how much of the fine-tuning leaks into the old task. The three parameters are swept jointly with the three taxonomical axes (Tab.~\ref{tab:grid}b), without assuming a one-to-one correspondence between them.

\textbf{Model mechanics.}
The model answers a question by generating a response of $T$ tokens, one at a time, from a vocabulary of size $K$, as an LLM does. At each position $t$ it has written a prefix $s=\bar y_{<t}$ of the response and picks the next token, so the prefixes form a finite tree, the empty prefix at the root and one branch per token\footnote{The reference setting uses $T=4$, $K=8$ and hidden dimension $D=64$, giving $\sum_{t=1}^{T}K^{t-1}=1+8+64+512=585$ nonterminal prefixes, including the empty prefix.}. Each prefix has a hidden state $\phi(s)\in\mathbb R^D$, drawn once at random and never updated, the analogue of the final hidden state at that position. The head $W\in\mathbb R^{K\times D}$ turns a hidden state into a next-token policy, $\operatorname{softmax}(W\phi(s))$. The same $W$ serves every prefix, so an update made at one prefix moves the policy at all others.

The old and new tasks ask different input ``questions'' and share the tree of prefixes, that is, the possible ``answers''. 
Unlike a transformer, which computes the hidden state at each response position from the question tokens and the response prefix, our model has no question tokens: the question is encoded directly in the hidden state.
The same prefix therefore has one hidden state under each task, $\phi_{\rm old}(s)$ and $\phi_{\rm new}(s)$, the new one built from the old so that their cosine is $\rho_\phi$ at every prefix. A change that moves the new-task policy at a prefix thus moves the old-task policy at the same prefix, the more so the larger $\rho_\phi$: at $\rho_\phi=1$ the two tasks share the hidden state and differ only in the answer, at $\rho_\phi=0$ their hidden states are orthogonal.

The pretrained model is a head $W_0$ with centered Gaussian entries, multiplied by $\lambda>0$. Scaling this head sharpens every initial policy without changing which token each prefix prefers, so $\lambda$ is how confidently the initial model holds its pretrained behavior: near zero, the policies are close to uniform and there is little to override; for large $\lambda$, each prefix has a firm preferred token. $W_0$ sets the two starting behaviors: $r_{\rm old}(\cdot\mid s)=\operatorname{softmax}(W_0\phi_{\rm old}(s))$ on the old task, the behavior to preserve, the analogue of the capabilities that retention measures in Sec.~\ref{sec:experimental_design}; and $p_0(\cdot\mid s)=\operatorname{softmax}(W_0\phi_{\rm new}(s))$ on the new task, the student's starting point. 

The new task is defined by hidden states $\phi_{\rm new}$ and its target, a next-token policy $r_{\rm new}$ at each prefix that the student must learn. For our model, we want the target to (i) prefer at least two tokens at each prefix, so that the mode-seeking and mode-covering KL (Sec.~\ref{sec:taxonomy}) can behave differently on it, and (ii) prefer tokens unrelated to the pretrained ones, so that the task can contradict a confident model. To do so, a second random head, orthogonal to $W_0$, ranks the tokens at each prefix, and the two top-ranked ones receive most of the mass (App.~\ref{rho_W_ablation} varies this alignment). 
Since we chose these distributions prefix by prefix, no single head can produce all of them at once: we therefore define $r_{\rm new}$ as the policy of the head $W_{\rm ref}$ that comes closest to them (App.~\ref{app:toy:construction}). By construction, the student can represent the target exactly.

The teacher is a copy of the student with privileged context: like any input the context acts only through the hidden state, so at optimization step $i$ the teacher reads a shifted hidden state through its own head $W_i^\star$,
$q_i(\cdot\mid s)=\operatorname{softmax}\!\left(W_i^\star[\phi_{\rm new}(s)+\psi_\kappa(s)]\right)$, while the student reads $p_i(\cdot\mid s)=\operatorname{softmax}(W_i\phi_{\rm new}(s))$. The shift $\psi_\kappa(s)$ is fixed before training and chosen, prefix by prefix, so that, with $W_0^\star=W_0$, the initial teacher's policy is exactly the mixture $q_0=(1-\kappa)p_0+\kappa r_{\rm new}$. The context strength $\kappa\in[0,1]$ is thus how much the context alone already helps the initial model: zero gives no initial teacher advantage, one gives the desired policy. The shift is fixed once and later read through the moving teacher head, as the context text is unchanged while the LLM reading it evolves; the model leaves out any learning in the hidden states, including how the context is encoded.

\textbf{Training and evaluation.} 
Training follows Sec.~\ref{sec:taxonomy}: rollouts from the student or the teacher, where instead of sampling responses we weight each prefix by its probability of being reached under the chosen policy, which the finite tree makes exact; forward or reverse KL at each visited prefix; an EMA of $W_i$ into $W_i^\star$ at rate $\alpha$. Two policies are compared by their similarity, one minus their total variation distance. Acquisition, the analogue of the accuracy gain, is the similarity of the final student to $r_{\rm new}$ on the prefixes it visits, minus the same quantity for the initial student on the prefixes it visits: each student is scored on the responses it would produce, as accuracy is. Retention, the analogue of the benchmark change, is the similarity of the final head, read on the old hidden states, to $r_{\rm old}$ on the prefixes $r_{\rm old}$ visits, minus its initial value of one; it is negative when the model forgets. Unless stated, figures use $\lambda=2.5$, a confident initial model, $\kappa \in \lbrace 0.3, 0.6, 0.9 \rbrace$ and $\rho_\phi=0.5$, within the grid of Tab.~\ref{tab:grid}b. Full construction, calibration and metrics are in App.~\ref{app:toy:backbone}--\ref{app:toy:metrics}.

%% file: llm_results/experimental_results.tex
\section{Experimental Results}
\label{sec:llm_results}

We now investigate the three axes, by (i) contrasting the effect on a matched pair of LLM configurations, (ii) assessing statistical relevance by a Wilcoxon test over the whole grid (Sec.~\ref{sec:experimental_design}),  and (iii) reproducing the pattern by sweeping  one factor of the controlled model (Sec.~\ref{sec:toy_model}).

\begin{figure}[t]
    \centering
    \begin{subfigure}[b]{0.575\linewidth}
        \centering
        \includegraphics[scale=1]{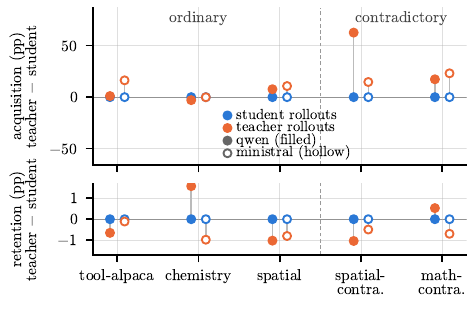}
        \vspace{-0.22cm}
        \caption{LLM experiments}
        \label{fig:rollout:exp}
    \end{subfigure}\hfill
    \begin{subfigure}[b]{0.415\linewidth}
        \centering
        \includegraphics[scale=1]{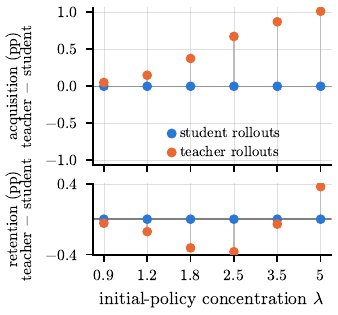}
        \vspace{-0.22cm}
        \caption{Controlled model}
        \label{fig:rollout:model}
    \end{subfigure}
    \vspace{-0.15cm}
    \caption{\textbf{Teacher rollouts raise acquisition where the pretrained behavior resists.} Paired difference teacher $\!-\!$ student ($>0$: teacher rollouts do better); the student is anchored at 0, the orange dot marks the difference. Same epochs and lr (Sec.~\ref{sec:experimental_design}), forward KL, each source at its best coupling $\alpha^{*}$. (a)~One pair per task and model. (b)~One pair per initial-policy concentration $\lambda$ (teacher advantage grows with $\lambda$), $\kappa=0.3$.}
    \label{fig:llm:rollout}
    \vspace{-0.5cm}
\end{figure}

\subsection{Rollout Source}%
\label{sec:llm_results_rollout_src}

Fig.~\ref{fig:llm:rollout}a compares the two rollout sources, each at its best coupling rate, under forward KL. On acquisition, teacher rollouts win every pair of the contradictory tasks by a wide margin. On the ordinary tasks the gap is small, and on some pairs the student wins. The Wilcoxon test over the whole grid confirms this beyond the best-coupling pairs: teacher rollouts add a median $+8$ acquisition points on the contradictory tasks for both models, and little or nothing on the ordinary ones, where one task even favors the student (Tab.~\ref{tab:app:stats-rollout}). Reverse KL gives the same result with similar gaps (App.~\ref{app:llm_results_rollout_src}). On retention, on the other hand, the rollout source makes almost no difference: the median gap stays under one point on both task types. Teacher rollouts thus matter where the task contradicts the pretrained behavior, as Sec.~\ref{sec:taxonomy} anticipated: the student's own responses stay close to what it already does, while the teacher's reach the prefixes where the new behavior is needed.

To vary how much the task contradicts the pretrained behavior, the controlled model (Fig.~\ref{fig:llm:rollout}b) varies how confident the initial model is, $\lambda$ (App.~\ref{app:teacher_utility_bridge} checks this correspondence on the spatial pair). Teacher rollouts win, and win more as the initial model grows more confident, with no clear retention preference. 
The model also shows where this advantage can stop: at the largest tested $\lambda$ and $\kappa$ (App.~\ref{app:rollout_transfer_boundary}), student rollouts acquire slightly more on average (this is because student rollouts follow similar trajectories to the ones used in test time).
Given the overall advantage of teacher rollouts in our experiments, the two remaining axes (Secs.~\ref{sec:llm_results_ema}--~\ref{sec:llm_results_kl_dir}) are studied with this source; 
their student counterparts
are in App.~\ref{app:ablations}.

\subsection{Teacher Coupling}\label{sec:llm_results_ema}

\begin{figure}[t]
    \centering
    \begin{subfigure}[b]{0.49\linewidth}
        \centering
        \includegraphics[scale=1]{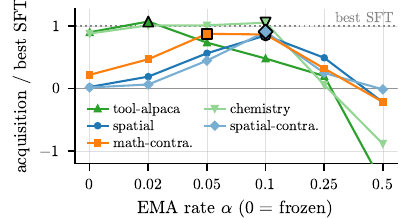}
        \vspace{-0.45cm}
        \caption{LLM experiments}
        \label{fig:coupling:exp}
    \end{subfigure}\hfill
    \begin{subfigure}[b]{0.49\linewidth}
        \centering
        \includegraphics[scale=1]{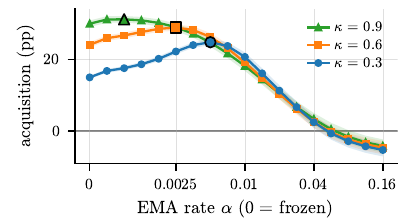}
        \vspace{-0.45cm}
        \caption{Controlled model}
        \label{fig:coupling:model}
    \end{subfigure}
    \vspace{-0.15cm}
    \caption{\textbf{A moving teacher helps up to a task-dependent rate, then hurts.}
    (a)~Qwen with teacher rollouts and forward KL, one learning rate (2e-5) and the same number of steps for every task. Acquisition is divided by the best SFT gain on the same task (dotted line: SFT; $0$: untrained accuracy; $-1$: a loss as large as SFT's gain). Black-edged markers: the best rate $\alpha^\star$.
    (b)~Acquisition against $\alpha$ for three levels of context strength $\kappa$ (shading: 95\% CI over 96 seeds).}
    \label{fig:llm:coupling}
    \vspace{-0.5cm}
\end{figure}

Having fixed teacher rollouts, we now vary the coupling rate $\alpha$, at the learning rate of Sec.~\ref{sec:experimental_design} and the same number of steps for every task.
Fig.~\ref{fig:llm:coupling}a shows forward KL on Qwen, with acquisition divided by the task's best SFT gain so that the curves are comparable across tasks (other cases in App.~\ref{app:llm_results_ema}). Every curve rises and then falls, but the peak depends on the task: for tool-alpaca and chemistry a frozen teacher is nearly the best, whereas math-contradiction and the two spatial tasks, one of them ordinary, gain most of their acquisition from a moving one. Past the peak every task falls, and at the fastest rate all are at or below their untrained accuracy. The Wilcoxon test over the whole grid (Tab.~\ref{tab:app:stats-coupling}) confirms this and adds the model: on Qwen a moving teacher beats the frozen one at every rate up to $0.10$, for a median retention cost below half a point; on Ministral the gain peaks earlier, at $0.05$, and turns into a loss at $0.25$, with a growing retention cost. The frozen-versus-moving disagreement between SDFT~\citep{shenfeld2026sdft}, SDPO~\citep{hubotter2026sdpo} and \citet{kim2026degrade} can thus come from where each study's rate falls: a moving teacher helps up to a rate that depends on the task,  the model  and the privileged context (App.~\ref{app:llm_context_ablations}), and hurts beyond it.

The controlled model (Fig.~\ref{fig:llm:coupling}b) varies the context strength $\kappa$, how far the context alone moves the initial model toward the new task's answers (Sec.~\ref{sec:toy_model}). The LLM runs do not measure $\kappa$; its closest indication is the gain of the frozen teacher in Fig.~\ref{fig:llm:coupling}a, which ranges from a few percent of the SFT gain to nearly all of it across tasks. Acquisition again rises and then falls with $\alpha$ for every $\kappa$. As $\kappa$ grows, the peak moves to slower rates and the gain over a frozen teacher shrinks: when the context already helps a lot, moving the teacher adds little. The same holds under reverse KL with student rollouts, and a corrupted context can even make the frozen teacher the best (App.~\ref{app:coupling_controls}).

\begin{figure}[t]
    \centering
    \begin{subfigure}[b]{0.655\linewidth}
        \centering
        \includegraphics[scale=1]{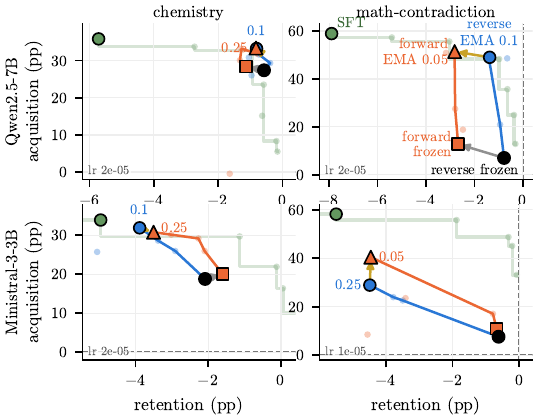}
        \vspace{-0.55cm}
        \caption{LLM experiments}
    \end{subfigure}\hfill
    \begin{subfigure}[b]{0.327\linewidth}
        \centering
        \includegraphics[scale=1]{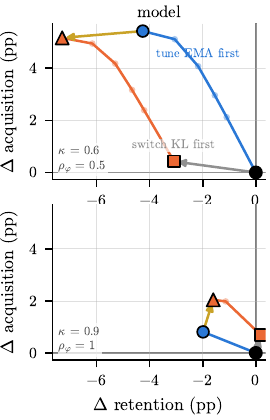}
        \vspace{-0.55cm}
        \caption{Controlled model}
    \end{subfigure}
    \vspace{-0.22cm}
    \caption{\textbf{Coupling and KL direction incur different retention costs.}
(a)~Teacher rollouts, one lr per panel (corner note; the one at which frozen reverse KL acquires most). Routes from frozen reverse KL: the EMA coupling tuned under each direction (blue: reverse; orange: forward; best $\alpha$ labeled), the direction switch frozen (gray) and between the optima (gold). Green: the SFT grid and its Pareto front.
(b)~The model counterpart ($\lambda=2.5$, $\kappa$ and $\rho_\phi$ in the corner notes), axes relative to frozen reverse KL.}
    \label{fig:llm:axis-tradeoffs}
    \vspace{-0.5cm}
\end{figure}

\subsection{Token-Level Divergence}\label{sec:llm_results_kl_dir}

We finally vary the KL direction and weight it against coupling in Fig.~\ref{fig:llm:axis-tradeoffs}a.
Each panel starts from frozen reverse KL (the setting that keeps training closest to the pretrained model) and compares the two ways of moving away from it: tuning the coupling rate under reverse KL 
(blue route), or first switching to forward KL (gray arrow) and then tuning the coupling there (orange route), at the learning rate (out of three) where 
frozen reverse KL acquires most.
On Qwen, tuning the coupling rate under reverse KL buys most of the acquisition for a small retention loss ($+42.0$ points for $0.6$ on math-contradiction), whereas switching to forward KL adds far less acquisition than tuning the coupling rate, for a larger retention loss: $+5.8$ for $1.9$ with a frozen teacher, and  $+2.4$ for $1.5$ once the coupling rate is tuned on each side (gold arrow).
Two new seeds and a separate test split lead to the same conclusion (App.~\ref{app:llm_results_test_split}).
On Ministral, the better order is the other one: switching to forward KL costs no retention on either task, and on math-contradiction the forward optimum adds $11.5$ points over the reverse one at the same retention. A third model family, Gemma-4-12B, behaves like Qwen on math-contradiction (App.~\ref{app:llm_results_gemma}).
Which move to make first thus appears to be \textit{model-dependent}. 
The Wilcoxon test over the whole grid (Tab.~\ref{tab:app:stats-kl}) confirms this beyond the two tasks shown: on Qwen, forward KL reaches the highest acquisition on every task, but reverse KL retains more in nearly every pair, by half a point on median; on Ministral, forward KL retains more when the pairs are pooled, and improves both metrics at once in a third of them. Pooled over the grid, the KL direction thus moves acquisition least of the three axes. At the best coupling rate it can add far more, as for Ministral math-contradiction. Its retention cost depends on the model: the forward KL preferred by SDFT~\citep{shenfeld2026sdft} costs some retention on Qwen and none on Ministral.
Finally, the controlled model reproduces this dependence (Fig.~\ref{fig:llm:axis-tradeoffs}b): raising the context strength and the feature overlap turns the Qwen behavior into the Ministral one. This change holds over a contiguous region of the grid, not only at the two settings plotted (App.~\ref{app:kl_regimes}).

SFT trained on the same demonstrations, over a wider learning-rate grid, traces the green Pareto front of Fig.~\ref{fig:llm:axis-tradeoffs}a. Most self-distillation families end above that front 
($17$ of $20$ on Qwen, $14$ of $20$ on Ministral using the fixed learning rate described in Sec.~\ref{sec:experimental_design}, Tab.~\ref{tab:app:sft-crossing}; $17$ and $11$ when the self-distillation learning rate is also selected, Tab.~\ref{tab:app:sft-crossing-lr-free}):
at the acquisition they reach, they retain more than SFT. Almost none reaches the highest acquisition, however: SFT keeps that lead, at $3$ to $21$ points of retention, everywhere except Qwen on tool-alpaca and chemistry. The claim that SDFT ``consistently outperforms SFT''~\citep{shenfeld2026sdft} thus holds for the trade-off, not for each metric separately.

%% file: conclusion/conclusion.tex
\section{Conclusion}
\label{sec:conclusion}

By writing self-distillation methods as one objective with three axes, and sweeping every combination of the three on two LLMs and five tasks with SFT as the reference, 
we help reconcile conclusions that the literature reports as conflicting. A frozen teacher and a moving one, reported as better and as worse, can each be the better one, depending on the task and on the coupling rate each study uses.
Teacher and student rollouts differ mostly where the task contradicts the pretrained behavior. The retention cost of the KL direction, found in one study and absent in another, depends on the model. A controlled model reproduces these trends by varying three quantities: initial-policy concentration, context strength, and feature overlap. The right setting thus depends on the task and the model, and the three choices are best evaluated together. 
Our experiments cover five tasks, two model families (a third on two tasks, App.~\ref{app:llm_results_gemma}) and two seeds per configuration (checked with new seeds, App.~\ref{app:llm_results_seeds}), with the reference response as context and other contexts (a hint and feedback) only as ablations (App.~\ref{app:llm_context_ablations}).
We do not measure the three quantities of the controlled model on the LLMs. Estimating them from a model and a task before training, and extending the controlled model to contexts that change during training, such as feedback, are left to future work.

%% file: appendix/appendix.tex
\appendix

\section{Per-Row Justification for Tab.~\ref{tab:related}}
\label{app:detailed_related_work}

In Tab.~\ref{tab:related}, we included a setting whenever the study reports the corresponding results in the main paper or as an ablation. 
Axes marked in bold are linked with paper reporting the whole combinations over those axes.
When the trajectories are offline (using a hard-target loss) and following a frozen teacher, e.g. for \citet{yang2024sdft}, we wrote the rollout source as ``teacher'', even if it is not a rollout source of its own in that case.

\paragraph{Distillation from a fixed external teacher.}
GKD~\citep{agarwal2024gkd} trains on a mixture of student generations and responses supplied in the training dataset, sweeping the student fraction over $0$, $0.5$ and $1$, and compares forward KL, reverse KL and three JSD variants, reporting the two axes on one grid; \emph{student/data} denotes these two response sources, since the dataset responses need not be teacher-generated. Decoupling KL~\citep{zhao2026decoupling} runs all four combinations of student or teacher prefixes with forward or reverse KL as named configurations, and its appendix scores the initial student and all four distilled models on a separate capability suite. Prefix OPD~\citep{zhang2026prefixopd} crosses the same two axes at their endpoints under a $256$-token training prefix, its off-policy prefixes being fixed corpus trajectories generated by a third model. In all three the teacher is larger and fixed, so teacher coupling does not arise.

\paragraph{Self-teacher without privileged context.}
\citep{guo2026teachermovetemporalcoupling} vary the coupling schedule alone, comparing a fixed teacher, periodic hard refreshes, two EMA rates and an adaptive rule, under a reinforcement objective whose distillation term is implemented as a generalized JSD despite the reverse-KL notation of its setup. Their teacher is a historical copy of the student and carries no privileged context, so the row is listed separately, and their held-out accuracies concern the trained tasks themselves.

\paragraph{Self-teacher conditioned on privileged context.} 

\citet{yang2024sdft} let the initial model rewrite each reference response while seeing it and train on the rewrites with cross-entropy, so the rollouts are the teacher's and no divergence direction is defined, hence \emph{hard targets}.
SDFT~\citep{shenfeld2026sdft} samples from the student and uses forward KL in practice while motivating reverse KL in theory, reporting the latter as worse without a direction-resolved ablation; its appendix compares a frozen teacher, the current student and an EMA teacher with three different coupling rates.
OPSD~\citep{zhao2026opsd} fixes the teacher to the initial policy in its experiments, and compares forward KL, reverse KL and JSD, forward performing best. 
SDPO~\citep{hubotter2026sdpo} uses a symmetric JSD on student rollouts without comparing directions, and compares four teachers, including: frozen, current, EMA at rate $0.01$, and a trust-region interpolation.
OPCD~\citep{ye2026opcd} pairs student rollouts and reverse KL against a context-distillation baseline pairing teacher rollouts with forward KL, two opposite corners of a crossing and not the crossing itself, and separately compares a frozen teacher with a continuously updated self-teacher.
Skill-SD~\citep{wang2026skillsd} reports all four combinations of student or teacher rollouts with a frozen or synchronized teacher, the only rollout-by-coupling crossing among self-distillation studies, both teacher-rollout runs collapsing mid-training. 
\citet{kim2026degrade} study an offline cross-entropy setting and an online JSD setting, whose rollout sources change together with the objective, and compare a teacher fixed at the initial policy with one updated at rate $0.05$, finding the fixed teacher better.
DiSC~\citep{padmanabhan2026disc} minimizes a forward KL between a frozen document-conditioned teacher and the student on fixed document suffixes under teacher forcing, which are corpus tokens and not generated trajectories, hence \emph{data}. 
MixSD~\citep{liu2026mixsd} draws each target token from a Bernoulli mixture of the same frozen model with and without the injected knowledge, so neither branch is the updating student, and trains with cross-entropy on the mixture; we therefore mark the source as \emph{teacher+base}, corresponding respectively to the knowledge-conditioned and unconditioned branches. 
GATES~\citep{stein2026gates} uses consensus-gated tutor trajectories and student trajectories together; its tutor shares weights with the student, and its on-policy term weights student likelihoods by a clipped teacher--student log-ratio, corresponding to a clipped reverse-KL-style update and never compared against forward KL.

\section{Datasets}\label{app:datasets_contexts}

We describe the datasets related to the five tasks considered in the paper. Following Sec.~\ref{sec:taxonomy}, each training example contains a task input $x$ and a reference response $y^{\mathrm{ref}}$. 
The input $x$ contains the question and additional information (tool documentation for tool-alpaca, answer options for chemistry, the spatial relation between the entities for spatial and spatial-contradiction). The reference response $y^{\mathrm{ref}}$ contains an explanation of the solution followed by the reference answer. The student receives only $x$. The teacher receives the privileged context $c=y^{\mathrm{ref}}$ inserted into a fixed task-specific template. For SFT, we use $y^{\mathrm{ref}}$ as the response target.
After training, we evaluate the student without privileged context on each task's held-out validation set. The splits are summarized in Tab.~\ref{tab:task-baselines}.

\begin{table*}[h]
\centering
\small
\caption{Training and validation set sizes and initial task accuracies $a_0$
(\%). The spatial pair shares all inputs and split assignments, changing only
the reference responses and answers.}
\label{tab:task-baselines}
\begin{tabular}{@{}lrrrr@{}}
\toprule
 & & & \multicolumn{2}{c}{Initial accuracy $a_0$ (\%)} \\
\cmidrule(l){4-5}
Task & $N_{\rm train}$ & $N_{\rm val}$ & Qwen2.5-7B & Ministral~3~3B \\
\midrule
tool-alpaca & 4,046 & 97 & 41.20 & 6.53 \\
chemistry & 2,674 & 507 & 33.50 & 36.36 \\
spatial & 4,000 & 500 & 26.20 & 19.60 \\
spatial-contradiction & 4,000 & 500 & 1.50 & 0.80 \\
math-contradiction$^\dagger$ & 4,000 & 500 & 13.45 & 12.40 \\
\bottomrule
\end{tabular}

\vspace{2pt}
\begin{minipage}{0.96\textwidth}
\footnotesize
$^\dagger$For comparison, initial accuracy on the corresponding non-contradicted
Math dataset is 91.4\% for Qwen2.5-7B and 82.4\% for Ministral~3~3B.
\vspace{-0.4cm}
\end{minipage}
\end{table*}

In this section, we describe how each dataset was built (or, for tool-alpaca and chemistry, where it comes from) and how answers are scored.

\subsection{ToolAlpaca and SciKnowEval Chemistry L-3}\label{app:llm:ordinary-datasets}

\paragraph{Provenance and supervision.}

We use the ToolAlpaca~\citep{tang2023toolalpaca} (tool-alpaca) and SciKnowEval Chemistry L-3~\citep{feng2024sciknoweval} (chemistry) training and evaluation sets and reference responses used by SDFT~\citep{shenfeld2026sdft}. Tool-alpaca gives a user request and available tool documentation; its reference response contains the intended action and JSON arguments. Chemistry asks a four-option question and provides a reference explanation ending in an option label. Tool-alpaca's demonstrations come from the source dataset. For chemistry, SDFT generated up to 8 GPT-4o responses per question and retained one with the correct final option. In tool-alpaca, both models see the user request and tool documentation; in chemistry, both see the question and its answer options. Only the full reference response is reserved for the teacher and used as the SFT target.

\paragraph{Scoring.}
Tool-alpaca receives a binary score when both the multiset of predicted action names and the JSON argument dictionary formed by merging arguments across calls match the reference. Call order and JSON key order do not affect the score, but action multiplicity and argument values do. This structural metric does not execute tools or check which arguments belong to which action in a response with multiple calls. Chemistry uses case-sensitive exact match between the extracted \texttt{<answer>} option and the reference label. Neither task scores the accompanying explanation; in particular, selecting a correct chemistry option does not establish that its reference reasoning is sound.

\subsection{Math Contradiction}
\label{app:llm:math-construction}

We introduce math-contradiction, a new dataset built on the DeepMind Mathematics Dataset~\citep{saxton2019mathematics}. 
It changes the numeral convention while preserving ordinary arithmetic and algebraic operations. We draw 10,000 source questions from six modules of the DeepMind Mathematics Dataset: mixed arithmetic (1,668), repeated addition/subtraction (1,668), repeated multiplication/division (1,668), division (1,666), and linear equations in one unknown (1,665) and two unknowns (1,665). Source sampling is approximately balanced across these modules and the easy/medium/hard difficulties. We retain questions with numeric answers for automated verification.

\paragraph{Reference construction and checks.}
For each source problem, Qwen3-30B-A3B-Instruct-2507 generates two step-by-step candidate solutions. We retain the shorter candidate by token count among those that complete normally, recover the correct source answer in a single final box, and satisfy the numeral-conversion checks. This yields one reference response for each of 8,432 source problems. We then rewrite numeric literals in the question, reference answer, and retained solution in base nine, preserving their values and the underlying operations. When an accepted value has no finite base-nine expansion, it is represented by an exact fraction: for example, $0.5_{10}=0.\overline{4}_9$ is written as $1/2$, with both integers interpreted in base nine. Unsupported numeric notation, such as scientific notation written as \texttt{1e3}, is rejected. After this process, the converted problems and responses all use ordinary arithmetic expressed in base-nine positional notation as an implicit convention.

After conversion, we check that the solution's boxed answer represents the same number as the original correct answer. For example, the decimal calculation $7+5=12$ becomes $7+5=13$ in base nine, where $13_9=12_{10}$. However, some questions have the same written answer under both conventions: $2+3$ gives $5$ in either base, while $13/1$ gives $13$ in either base. Such examples can be answered correctly without using the changed numeral convention. We remove 577 questions whose text is unchanged by conversion. A pattern-based filter removes another 577 examples involving division or multiplication by one, addition or subtraction of zero, or multiplication by zero, targeting cases where copying a numeral or returning zero can produce the answer.

For the remaining examples, we use SymPy to evaluate supported arithmetic expressions and solve linear equations in the transformed questions, interpreting their numerals as decimal. We discard an example if this computed answer matches the converted reference answer, also interpreted as decimal: ordinary decimal solving would then produce an accepted answer without requiring the base-nine convention. This check removes 953 further examples, leaving 6,325 for the subsequent full-solution audit. Questions that the symbolic solver cannot parse or solve are not removed by this check.

Because a correct final answer does not ensure that the converted explanation is coherent, GPT-5 mini separately audits the full solution for arithmetic, algebraic, and base-semantic consistency, including positional reasoning and carry/borrow operations. For example, an excluded solution to $-65823_9 / -126_9$ gives the correct answer $512_9$ but describes $534$ as the dividend's first three digits, which are actually $658$. Only examples judged consistent and recommended for retention enter the final pool: 5,772 are retained, while 532 are labeled inconsistent, 20 suspect, and one has an audit error. A deterministic shuffle of this pool 
assigns 4,000 examples to training, 500 to validation, and the remaining 1,272 to a test split used only in App.~\ref{app:llm_results_test_split}, keeping source-example identities disjoint.

\paragraph{Answer scoring.}
Evaluation reads the last complete boxed answer using base-nine semantics and compares its numeric value with the reference answer. For example, $10_9$ is interpreted as nine, not ten. Supported integer and rational answers are compared exactly, so equivalent fractions receive the same score. For expressions outside this exact numeric grammar, scoring falls back to Math-Verify after base-nine numerals have been converted to exact decimal representations. Answers without a complete box or with invalid base-nine digits are rejected. This check applies to the final answer; it does not grade the generated reasoning.

\paragraph{Example.}
The student prompt requests a boxed answer but never announces the numeral
base. In the example in Sec.~\ref{app:llm:example-math}, the steps
$-3-6=-10$ and $13+(-10)=3$ are valid when all numerals are read in base nine:
$10_9=9_{10}$ and $13_9=12_{10}$. The full expression has answer $3$ under this
convention, but $4$ under an ordinary decimal reading. Success therefore
requires transferring the implicit convention from training references to
unseen problems. The resulting accuracy measures this transfer within the
constructed distribution, rather than a general revision of arithmetic
behavior in other settings.

\subsection{Spatial Task, with and without Contradiction}
\label{app:llm:spatial-construction}

The spatial tasks ask for an entity's coordinates after composing a chain of
relations on a square grid. Each problem provides one anchored entity with a
known coordinate pair, the relations linking it to a queried entity, and the
query. The construction spans one to six relations and uses cardinal or diagonal
steps of one to four grid cells. Relations are presented in shuffled order, and
some are stated in the direction opposite to traversal from the anchor. Solving
the problem therefore requires ordering the links, reversing the relevant
displacements, and composing them. The number of links, or \emph{hops}, controls
compositional difficulty independently of the direction convention.

\paragraph{Construction and language variation.}
We first generate a symbolic chain of distinct entities at distinct integer grid coordinates, then render its relations and step-by-step solutions in natural language. Each split contains approximately equal numbers of one- to six-hop problems. During chain generation, directions are proposed uniformly from the four cardinal and four diagonal directions, and step lengths of one to four cells are proposed with probabilities $(0.48,0.30,0.16,0.06)$; proposals that revisit an occupied coordinate are resampled. Each relation is stated in reverse with probability $0.35$. Anchors are sampled at the origin with probability $0.20$ and otherwise uniformly from the nonzero integer points in $[-4,4]^2$. A $k$-cell diagonal means $k$ diagonal grid steps, changing both coordinate magnitudes by $k$.The wording varies across compass terms, relative-position descriptions such as ``above'' or ``to the right,'' clock-face directions, compass bearings, and instructions involving turns. Phrase selection respects grammatical compatibility with the surrounding sentence.

Both reference responses are computed from the symbolic chain and rendered with matched wording choices. We generate 4,000 training examples, 500 validation examples, and 2,000 examples in a separate test split. Construction checks recompute the coordinates, verify the final answers, and reject repeated problem texts or identical named relation chains across the generated splits. All 6,500 examples pass these checks, with no repeated problem texts or named relation chains across splits. These checks permit shared reasoning patterns across distinct examples. We additionally used ChatGPT 5.4 to inspect rows for potentially unnatural constructions, followed by manual inspection of samples from every wording family. These language-quality checks complement the symbolic checks; they are not used to score model predictions.

\paragraph{Ordinary and rotated semantics.}
Let $\mathbf v_e$ be a relation's displacement under the ordinary convention,
with east $(1,0)$ and north $(0,1)$. Spatial-contradiction uses a fixed clockwise
rotation of each displacement:
\begin{equation}
\mathbf v_e^{\rm rot}=R\mathbf v_e,\qquad
R=\begin{pmatrix}0&1\\-1&0\end{pmatrix}.
\label{eq:app:spatial-rotation}
\end{equation}
Thus a unit northward step contributes $(1,0)$, an eastward step $(0,-1)$, and
a northeast step $(1,-1)$. A turn description first determines a direction,
which is then interpreted using the same convention. The rotation is fixed
across all links; it does not accumulate with hop count. For anchor coordinate
$\mathbf a$ and path $P$, the targets are
\begin{equation}
\mathbf z_{\rm ordinary}=\mathbf a+\sum_{e\in P}\epsilon_e\mathbf v_e,
\qquad
\mathbf z_{\rm rot}=\mathbf a+R\sum_{e\in P}\epsilon_e\mathbf v_e,
\label{eq:app:spatial-targets}
\end{equation}
where $\epsilon_e\in\{-1,+1\}$ indicates whether the relation is traversed
against or along its stated direction. All spatial vectors are columns.
The numerical anchor and output coordinate frame remain fixed; only the
relation displacements change.

\paragraph{Paired control and scoring.}
Spatial and spatial-contradiction use exactly the same task inputs $x$ and split assignments: entity names, anchors, relations, wording, and presentation order agree example by example. They differ only in the reference responses $y^{\mathrm{ref}}$ and the answers used for scoring. The two versions therefore have identical hop difficulty. The generated chains never return to the anchor, so the net displacement is nonzero and the two reference answers differ for every example. The prompt requests a boxed coordinate pair without revealing the alternative direction convention. Both tasks score the last complete boxed integer pair by exact coordinate equality with the appropriate reference answer. Missing boxes and non-integer pairs are rejected; the explanatory text is not scored.

\paragraph{Example.}
The example in Sec.~\ref{app:llm:example-spatial} has anchor $(-1,-1)$ and
ordinary net displacement $(5,1)$, giving target $(4,0)$. Under the rotated
convention, the displacement becomes $(1,-5)$ and the target is $(0,-6)$.
Both solutions traverse the same four links, including a final reversed link.

\section{Student Prompts, Reference Responses, and Teacher Templates}\label{app:llm:templates}

The examples below show how a task input $x$ and its full reference response
$y^{\mathrm{ref}}$ become student and teacher prompts. Chemistry uses a system
message followed by a user message; the other four tasks use a single user
message. The student receives the original task messages unchanged. The teacher
preserves the same role structure, but its user message contains both $x$ and the
privileged reference response. In the notation of Sec.~\ref{sec:taxonomy},
$c=y^{\mathrm{ref}}$; the template determines how this reference is presented. The reference is included as text inside the teacher input, not shown to the student and not inserted as an earlier assistant turn.
This section reports teacher prompts (App.~\ref{prompts:teacher})  and provides examples of the student messages and complete reference response for each task (App.~\ref{example:first}--\ref{example:last}). For all verbatim text, long lines are wrapped for readability. 

\subsection{Teacher prompts}\label{prompts:teacher}

As much as possible, templates are similar across tasks.
In particular, the same template is shared by tool-alpaca and chemistry: chemistry retains its system message; tool-alpaca has no additional system
message.  A slightly different template is shared by Math contradiction and both spatial tasks (contradictory and not contradictory).

For readability, the templates below use the explanatory placeholders \texttt{<TASK INPUT>} for the complete student-facing user content and \texttt{<REFERENCE RESPONSE>} for the full $y^{\mathrm{ref}}$. These markers are
not literal tokens in the training prompts. ``Fixed template'' means \textit{fixed within a task} across the compared configurations (not identical wording across all tasks).  Teacher templates are reported below.

\begin{tcolorbox}[width=\textwidth,colframe=black!75!white,title=Teacher user message (tool-alpaca and chemistry).]
\begin{Verbatim}[fontsize=\scriptsize,baselinestretch=0.9,breaklines,breakanywhere]
<TASK INPUT>

This is an example for a response to the question:
<REFERENCE RESPONSE>

Now answer with a response of your own, including the thinking process.
\end{Verbatim}
\end{tcolorbox}

\begin{tcolorbox}[width=\textwidth,colframe=black!75!white,title=Teacher user message (Math Contradiction and both spatial tasks).]
\begin{Verbatim}[fontsize=\scriptsize,baselinestretch=0.9,breaklines,breakanywhere]
<TASK INPUT>

Here is a verified reference solution for the same problem:
<REFERENCE RESPONSE>

Now solve the original problem yourself. Keep the requested final-answer format.
\end{Verbatim}
\end{tcolorbox}

The wording ``verified reference solution'' is part of the teacher template and should not be read as a claim that every natural-language intermediate step has been formally proved. SFT uses $y^{\mathrm{ref}}$ directly as the response trajectory, whereas self-distillation uses the token distribution of the reference-conditioned teacher on a sampled response $\bar y$. The two methods therefore receive the same reference information through different training objectives, as formalized in Sec.~\ref{sec:taxonomy}.

\subsection{Chemistry: Student Messages and Reference Response}\label{example:first}\label{app:llm:example-science}

\begin{tcolorbox}[width=\textwidth,colframe=black!75!white,title=Student system message (also retained by the teacher).]
\begin{Verbatim}

Given a question and four options, please select the right answer. Respond in the following format:
<reasoning>
...
</reasoning>
<answer>
...
</answer>

For the answer, only output the letter corresponding to the correct option (A, B, C, or D), and nothing else. Do not restate the answer text. For example, if the answer is "A", just output:
<answer>
A
</answer>
\end{Verbatim}
\end{tcolorbox}

\begin{tcolorbox}[width=\textwidth,colframe=black!75!white,title=Student user message.]
\begin{Verbatim}
What is the molar weight (g/mol) of the molecule with the the IUPAC name '(2S)-1-[(2R)-2-(benzenesulfonamido)-2-phenylacetyl]pyrrolidine-2-carboxamide'?

A: 427.900
B: 484.300
C: 387.500
D: 384.500
\end{Verbatim}
\end{tcolorbox}

\begin{tcolorbox}[splitbox,width=\textwidth,colframe=black!75!white,title=Reference response.]
\begin{Verbatim}
<reasoning>
To find the molar weight of the given molecule, we need to break down the IUPAC name and identify the atoms that constitute the molecule. 

1. The molecule is named '(2S)-1-[(2R)-2-(benzenesulfonamido)-2-phenylacetyl]pyrrolidine-2-carboxamide'.
2. The molecule contains the following key components:
   - Pyrrolidine: C4H9N
   - Carboxamide: CONH2
   - Phenylacetyl: C8H7O
   - Benzenesulfonamido: C6H5SO2NH

Next, we calculate the molar mass of each part and sum them:

- Pyrrolidine (C4H9N): 
  - C (Carbon): 12.01 * 4 = 48.04
  - H (Hydrogen): 1.01 * 9 = 9.09
  - N (Nitrogen): 14.01 * 1 = 14.01
  - Total: 71.14

- Carboxamide (CONH2):
  - C: 12.01
  - O (Oxygen): 16.00
  - N: 14.01
  - H: 1.01 * 2 = 2.02
  - Total: 44.04

- Phenylacetyl (C8H7O):
  - C: 12.01 * 8 = 96.08
  - H: 1.01 * 7 = 7.07
  - O: 16.00
  - Total: 119.15

- Benzenesulfonamido (C6H5SO2NH):
  - C: 12.01 * 6 = 72.06
  - H: 1.01 * 5 = 5.05
  - S (Sulfur): 32.07
  - O: 16.00 * 2 = 32.00
  - N: 14.01
  - Total: 155.19

Now sum up these molar masses: 
71.14 (pyrrolidine) + 44.04 (carboxamide) + 119.15 (phenylacetyl) + 155.19 (benzenesulfonamido) = 389.52 g/mol

Therefore, based on typical options provided, 387.500 g/mol is closest.

</reasoning>
<answer>
C
</answer>
\end{Verbatim}
\end{tcolorbox}

This reference selects option C after obtaining $389.52\,\mathrm{g/mol}$,
rather than C's $387.500\,\mathrm{g/mol}$, and choosing the closest option. We retain this discrepancy rather than silently correcting the training reference: it illustrates why final-label correctness does not certify the reasoning.

\subsection{Tool-alpaca: Student Message and Reference Response}
\label{app:llm:example-tooluse}

\begin{tcolorbox}[splitbox, width=\textwidth,colframe=black!75!white,title=Student user message (no system message).]
\begin{Verbatim}
Your task is to answer the user's question using available tools. 
You have access to the following tools:
Name: Httpbin
Description: A Simple HTTP Request & Response Service
Documentation:
sendHttpRequest: Send an HTTP request with the specified method, headers, and data to the Httpbin API for testing purposes.
Parameters: {"method": "Required. string. One of: [GET, POST, PUT, DELETE, HEAD, PATCH]. The HTTP method to use (GET, POST, PUT, DELETE, HEAD, or PATCH).", "url": "Required. string. The endpoint URL to send the request to.", "headers": "Object.  A key-value pair of headers to include in the request.", "data": "Object.  A key-value pair of data to include in the request body."}
Output: Successful response.
 - Format: application/json
 - Structure: Object{response: Object{status_code, headers: Object, body}}
getClientRequestData: Retrieve the client's request data, including headers, form data, uploaded files, and cookies.
Parameters: {"url": "Required. string. The endpoint URL to send the request to."}
Output: Successful response.
 - Format: application/json
 - Structure: Object{requestData: Object{headers: Object, form: Object, files: Object, cookies: Object}}
testProxyHeaders: Send a request to the Httpbin API's proxy headers testing endpoint.
Parameters: {"url": "Required. string. The endpoint URL to send the request to.", "headers": "Object.  A key-value pair of headers to include in the request."}
Output: Successful response.
 - Format: application/json
 - Structure: Object{response: Object{status_code, headers: Object, body}}
simulateStatusCode: Send a request to the Httpbin API's status code simulation endpoint to test how your application handles specific status codes.
Parameters: {"url": "Required. string. The endpoint URL to send the request to.", "statusCode": "Required. integer. The HTTP status code to simulate."}
Output: Successful response.
 - Format: application/json
 - Structure: Object{response: Object{status_code, headers: Object, body}}

Use the following format:
Thought: you should always think about what to do
Action: the action to take, should be one of the tool names.
Action Input: the input to the action, must be in JSON format. All of the action input must be realistic and from the user.

Begin!
Question: I need to test how my app handles a "PATCH" request. Can you send a PATCH request to https://httpbin.org/patch with the data '{"field": "value"}' and tell me the response?
\end{Verbatim}
\end{tcolorbox}

\begin{tcolorbox}[width=\textwidth,colframe=black!75!white,title=Reference response.]
\begin{Verbatim}
 I need to use the sendHttpRequest tool to send a PATCH request to the specified URL.
Action: sendHttpRequest
Action Input: {"method": "PATCH", "url": "https://httpbin.org/patch", "headers": {}, "data": {"field": "value"}}
\end{Verbatim}
\end{tcolorbox}

\subsection{Math Contradiction: Student Message and Reference Response}
\label{app:llm:example-math}

\begin{tcolorbox}[width=\textwidth,colframe=black!75!white,title=Student user message (no system message).]
\begin{Verbatim}
You will be given a problem.
Please reason step by step, and put your final answer within \boxed{}:
What is the value of (153 - 140) + (-3 - (-2 - -8))?
\end{Verbatim}
\end{tcolorbox}

\begin{tcolorbox}[splitbox, width=\textwidth,colframe=black!75!white,title=Reference response (base nine; not announced in the prompt).]
\begin{Verbatim}
Let's solve the expression step by step:

We are given:

$$
(153 - 140) + (-3 - (-2 - (-8)))
$$

---

### Step 1: Simplify inside the parentheses

Start with the first part:

$$
153 - 140 = 13
$$

Now simplify the innermost parentheses in the second part:

$$
-2 - (-8)
$$

Subtracting a negative is the same as adding:

$$
-2 + 8 = 6
$$

Now substitute back:

$$
-3 - (6) = -3 - 6 = -10
$$

---

### Step 2: Add the two results

$$
13 + (-10) = 13 - 10 = 3
$$

---

### Final Answer:

$$
\boxed{3}
$$
\end{Verbatim}
\end{tcolorbox}

\subsection{Spatial Tasks: Shared Student Message and Both Reference Responses}\label{example:last}
\label{app:llm:example-spatial}

\begin{tcolorbox}[width=\textwidth,colframe=black!75!white,title=Student user message; identical in both tasks (no system message).]
\begin{Verbatim}
You will be given a spatial reasoning problem on a square grid.
Distances are measured in grid cells. A k-cell diagonal relation means k diagonal grid steps.
Please reason step by step, and put your final coordinate pair within \boxed{(x, y)}:
On the grid, Bruno is at (-1, -1).

At the point one grid cell due north of Bruno, Boris is located.
To get from Farah to Faye, face west, make a quarter-turn clockwise, then move forward for two steps.
From Boris, Nia is 2 grid squares along the northeast diagonal.
Starting at Nia, face north, turn right by 90 degrees, then move forward for 3 grid cells to reach Faye.

Find the coordinate pair for Farah.
\end{Verbatim}
\end{tcolorbox}

\begin{tcolorbox}[width=\textwidth,colframe=black!75!white,title=Spatial reference response (ordinary semantics).]
\begin{Verbatim}
Take Bruno's coordinate, (-1, -1), as the fixed reference point.
Once the shuffled links are ordered, the traversal is Bruno -> Boris -> Nia -> Faye -> Farah.
Step 1: Read this relation in its given direction, taking us from Bruno to Boris.
Move one step due north from (-1, -1): add (0, +1) to obtain Boris at (-1, 0).
Step 2: No inversion is needed for the link Boris -> Nia.
Boris is at (-1, 0). Travel two cells toward the top right; applying (+2, +2) places Nia at (1, 2).
Step 3: The stated relation already points from Nia toward Faye, so use it as written.
From Nia at (1, 2), move three grid cells directly east. This gives displacement (+3, 0), so Faye is at (4, 2).
Step 4: The next connection is backward relative to the route, so reverse its direction to move from Faye to Farah.
The leg from Faye to Farah goes two squares straight south, corresponding to (0, -2). Numerically, (4, 2) + (0, -2) = (4, 0).
So the final coordinates of Farah are \boxed{(4, 0)}.
\end{Verbatim}
\end{tcolorbox}

\begin{tcolorbox}[width=\textwidth,colframe=black!75!white,title=Spatial Contradiction reference response (rotated semantics).]
\begin{Verbatim}
Take Bruno's coordinate, (-1, -1), as the fixed reference point.
Once the shuffled links are ordered, the traversal is Bruno -> Boris -> Nia -> Faye -> Farah.
Step 1: Read this relation in its given direction, taking us from Bruno to Boris.
Move one step due north from (-1, -1): add (+1, 0) to obtain Boris at (0, -1).
Step 2: No inversion is needed for the link Boris -> Nia.
Boris is at (0, -1). Travel two cells toward the top right; applying (+2, -2) places Nia at (2, -3).
Step 3: The stated relation already points from Nia toward Faye, so use it as written.
From Nia at (2, -3), move three grid cells directly east. This gives displacement (0, -3), so Faye is at (2, -6).
Step 4: The next connection is backward relative to the route, so reverse its direction to move from Faye to Farah.
The leg from Faye to Farah goes two squares straight south, corresponding to (-2, 0). Numerically, (2, -6) + (-2, 0) = (0, -6).
So the final coordinates of Farah are \boxed{(0, -6)}.
\end{Verbatim}
\end{tcolorbox}

\section{Additional Ablation Studies}\label{app:ablations}

This appendix first details the results of Sec.~\ref{sec:llm_results} on the full grid (App.~\ref{app:llm_results_main_grid}), then checks whether they hold beyond the main protocol: with acquisition measured at the final checkpoint (App.~\ref{app:llm_results_same_ckpt}), with new seeds and an unused test split (App.~\ref{app:llm_results_seeds}), with other privileged contexts (App.~\ref{app:llm_context_ablations}), and on a third model family (App.~\ref{app:llm_results_gemma}).

\subsection{Main-Grid Results}\label{app:llm_results_main_grid}
We detail the comparisons of Sec.~\ref{sec:llm_results} on the full grid, first axis by axis (Apps.~\ref{app:llm_results_rollout_src} to~\ref{app:llm_results_kl_dir}), then against SFT (App.~\ref{app:llm_results_sft}). For the axes, the main evidence comes from the Wilcoxon tests of Sec.~\ref{sec:experimental_design}: Tabs.~\ref{tab:app:stats-rollout}, \ref{tab:app:stats-coupling} and~\ref{tab:app:stats-kl} report them for rollout source (teacher $-$ student), coupling (EMA rate $\alpha$ against a frozen teacher, at a fixed number of steps) and token-level divergence (forward $-$ reverse). Each table compares, per model and task, matched pairs of grid configurations that differ only in the axis under study; the shared conventions are stated in the caption of Tab.~\ref{tab:app:stats-rollout}.

\subsubsection{Rollout Source}\label{app:llm_results_rollout_src} 

\begin{figure}[!b]
    \centering
    \begin{subfigure}[b]{0.575\linewidth}
        \centering
        \includegraphics[scale=1]{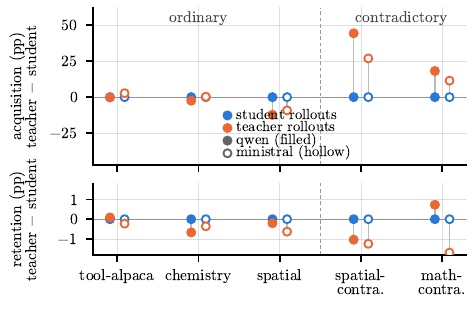}
        \vspace{-0.2cm}
        \caption{LLM experiments}
    \end{subfigure}\hfill
    \begin{subfigure}[b]{0.415\linewidth}
        \centering
        \includegraphics[scale=1]{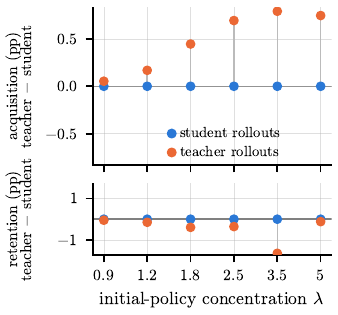}
        \vspace{-0.2cm}
        \caption{Controlled model}
    \end{subfigure}
    \vspace{-0.2cm}
    \caption{\textbf{Rollout-source axis.} Same as Fig.~\ref{fig:llm:rollout}, with reverse instead of forward KL.}
    \label{fig:app:llm:rollout_reverse}

    \vspace{1.2em}
    \setcounter{subfigure}{0}
    \centering
    \begin{subfigure}[b]{0.46\linewidth}
        \centering
        \includegraphics[width=\linewidth]{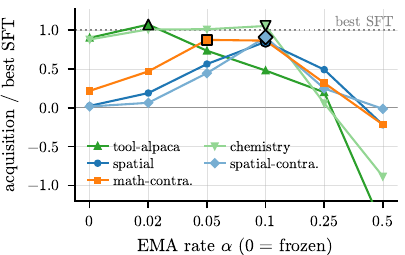}
        \caption{Qwen forward (same as Fig.~\ref{fig:llm:coupling}a)}
        \label{fig:axis2_qwen_forward}
    \end{subfigure}\hfill
    \begin{subfigure}[b]{0.46\linewidth}
        \centering
        \includegraphics[width=\linewidth]{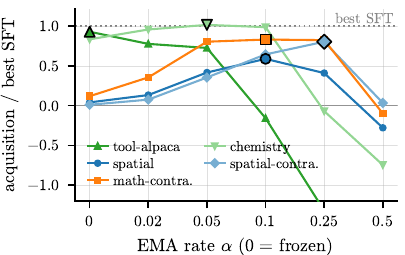}
        \caption{Qwen reverse}
        \label{fig:axis2_qwen_reverse}
    \end{subfigure}

    \vspace{0.5em}

    \begin{subfigure}[b]{0.46\linewidth}
        \centering
        \includegraphics[width=\linewidth]{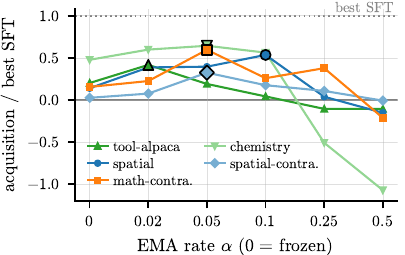}
        \caption{Ministral forward}
        \label{fig:axis2_ministral_forward}
    \end{subfigure}\hfill
    \begin{subfigure}[b]{0.46\linewidth}
        \centering
        \includegraphics[width=\linewidth]{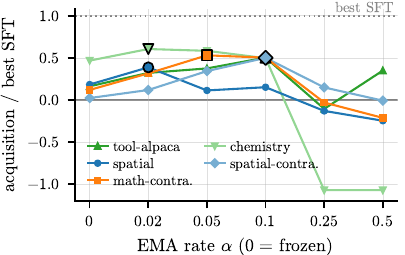}
        \caption{Ministral reverse}
        \label{fig:axis2_ministral_reverse}
    \end{subfigure}
    \vspace{-0.2cm}
    \caption{\textbf{Coupling axis.} Extension of Fig.~\ref{fig:llm:coupling}a to both KL directions, for Qwen (lr 2e-5, top) and Ministral (lr 5e-6, bottom), with teacher rollouts and the same number of steps for every task.}%
    \label{fig:axis2_appendix}
\end{figure}

Fig.~\ref{fig:app:llm:rollout_reverse} shows the same comparison as Fig.~\ref{fig:llm:rollout} with reverse instead of forward KL. The observations are similar to the forward KL case, with the same ordering on the contradictory tasks.

The per-task tests on matched pairs are reported in Tab.~\ref{tab:app:stats-rollout}.

\begin{table}[!tp]
\centering
\small
\begin{tabular}{l rr rr}
\toprule
& \multicolumn{2}{c}{Qwen2.5-7B} & \multicolumn{2}{c}{Ministral-3-3B} \\
\cmidrule(lr){2-3} \cmidrule(lr){4-5}
Task ($n$ pairs) & acquisition & retention & acquisition & retention \\
\midrule
tool-alpaca (30)           & $+1.8$*\phantom{**}   & $-0.0$\phantom{***} & $+1.7$\phantom{***} & $-0.2$*** \\
chemistry (30)             & $-2.9$***             & $+0.3$*\phantom{**} & $+2.0$**\phantom{*} & $-0.1$\phantom{***} \\
spatial (30)               & $+2.4$***             & $-0.0$\phantom{***} & $+2.5$\phantom{***} & $-0.6$*** \\
spatial-contradiction (30) & $+4.9$***             & $-0.1$\phantom{***} & $+4.4$***           & $-1.0$*** \\
math-contradiction (30)    & $+12.7$***            & $+0.1$\phantom{***} & $+10.9$***          & $-0.7$*\phantom{**} \\
\midrule
ordinary (90)              & $+0.5$\phantom{***}   & $+0.1$\phantom{***} & $+2.2$**\phantom{*} & $-0.3$*** \\
contradictory (60)         & $+8.0$***             & $-0.0$\phantom{***} & $+8.3$***           & $-0.9$*** \\
\midrule
all tasks (150)            & $+1.9$***             & $+0.0$\phantom{***} & $+3.2$***           & $-0.4$*** \\
\bottomrule
\end{tabular}
\caption{\textbf{Rollout-source axis}: median paired difference (teacher $\!-\!$ student, pp) over matched pairs identical except for the rollout source ($5$ couplings $\times$ $2$ KL directions $\times$ $3$ learning rates). Here and in Tabs.~\ref{tab:app:stats-coupling} and~\ref{tab:app:stats-kl}: two-sided Wilcoxon signed-rank tests, Holm-corrected across tasks within each model and metric ({*}~$p<0.05$, {**}~$p<0.01$, {***}~$p<0.001$); grouped rows pool the corresponding pairs.}
\label{tab:app:stats-rollout}
\vspace{2em}

\footnotesize
\begin{tabular}{ll rr rr}
\toprule
& & \multicolumn{2}{c}{Qwen2.5-7B} & \multicolumn{2}{c}{Ministral-3-3B} \\
\cmidrule(lr){3-4} \cmidrule(lr){5-6}
Task ($n$ pairs) & $\alpha$ & acquisition & retention & acquisition & retention \\
\midrule
tool-alpaca (12/12)          & $0.02$ & $+2.5$*\phantom{**}     & $-0.13$\phantom{***}    & $+3.4$\phantom{***}     & $+0.08$\phantom{***} \\
                             & $0.05$ & $-2.4$\phantom{***}     & $-0.47$*\phantom{**}    & $+5.8$\phantom{***}     & $-0.06$\phantom{***} \\
                             & $0.10$ & $-18.8$**\phantom{*}    & $-0.67$**\phantom{*}    & $+4.0$\phantom{***}     & $-0.63$**\phantom{*} \\
                             & $0.25$ & $-51.7$**\phantom{*}    & $-0.88$**\phantom{*}    & $-16.3$*\phantom{**}    & $-3.12$**\phantom{*} \\
\midrule
chemistry (12/12)            & $0.02$ & $+4.3$**\phantom{*}     & $-0.25$*\phantom{**}    & $+4.6$**\phantom{*}     & $-0.02$\phantom{***} \\
                             & $0.05$ & $+6.0$**\phantom{*}     & $-0.28$\phantom{***}    & $+5.8$**\phantom{*}     & $-0.36$*\phantom{**} \\
                             & $0.10$ & $+6.3$**\phantom{*}     & $-0.04$\phantom{***}    & $+3.0$\phantom{***}     & $-0.56$*\phantom{**} \\
                             & $0.25$ & $+6.3$\phantom{***}     & $+0.00$\phantom{***}    & $-42.4$**\phantom{*}    & $-1.20$**\phantom{*} \\
\midrule
spatial (12/12)              & $0.02$ & $+5.8$**\phantom{*}     & $-0.07$\phantom{***}    & $+21.7$**\phantom{*}    & $-0.25$\phantom{***} \\
                             & $0.05$ & $+10.4$*\phantom{**}    & $-0.15$\phantom{***}    & $+14.9$\phantom{***}    & $-1.04$*\phantom{**} \\
                             & $0.10$ & $+21.9$*\phantom{**}    & $-0.22$\phantom{***}    & $-6.4$\phantom{***}     & $-2.91$**\phantom{*} \\
                             & $0.25$ & $+11.9$\phantom{***}    & $-0.48$**\phantom{*}    & $-20.3$**\phantom{*}    & $-3.66$**\phantom{*} \\
\midrule
spatial-contradiction (12/12) & $0.02$ & $+1.9$**\phantom{*}     & $-0.14$\phantom{***}    & $+4.8$**\phantom{*}     & $-0.10$\phantom{***} \\
                             & $0.05$ & $+7.5$**\phantom{*}     & $-0.20$\phantom{***}    & $+20.6$**\phantom{*}    & $-0.76$*\phantom{**} \\
                             & $0.10$ & $+17.1$**\phantom{*}    & $-0.26$\phantom{***}    & $+15.1$*\phantom{**}    & $-1.73$**\phantom{*} \\
                             & $0.25$ & $+10.7$**\phantom{*}    & $-0.44$**\phantom{*}    & $+5.0$\phantom{***}     & $-3.05$**\phantom{*} \\
\midrule
math-contradiction (12/12)   & $0.02$ & $+3.9$**\phantom{*}     & $-0.09$\phantom{***}    & $+1.5$\phantom{***}     & $-0.12$\phantom{***} \\
                             & $0.05$ & $+9.0$**\phantom{*}     & $-0.23$*\phantom{**}    & $+7.6$\phantom{***}     & $-1.27$*\phantom{**} \\
                             & $0.10$ & $+18.4$**\phantom{*}    & $-0.28$\phantom{***}    & $+8.1$\phantom{***}     & $-2.18$**\phantom{*} \\
                             & $0.25$ & $+21.6$**\phantom{*}    & $-0.27$\phantom{***}    & $-8.6$\phantom{***}     & $-4.62$**\phantom{*} \\
\midrule
ordinary (36/36)             & $0.02$ & $+4.3$***               & $-0.12$***              & $+7.9$***               & $-0.06$\phantom{***} \\
                             & $0.05$ & $+3.6$**\phantom{*}     & $-0.27$***              & $+6.5$***               & $-0.39$*** \\
                             & $0.10$ & $+5.8$\phantom{***}     & $-0.28$***              & $+0.8$\phantom{***}     & $-0.76$*** \\
                             & $0.25$ & $+2.3$\phantom{***}     & $-0.50$***              & $-18.6$***              & $-2.15$*** \\
\midrule
contradictory (24/24)        & $0.02$ & $+3.1$***               & $-0.14$*\phantom{**}    & $+2.1$**\phantom{*}     & $-0.12$\phantom{***} \\
                             & $0.05$ & $+8.5$***               & $-0.23$**\phantom{*}    & $+14.4$***              & $-0.84$*** \\
                             & $0.10$ & $+18.4$***              & $-0.26$**\phantom{*}    & $+10.4$*\phantom{**}    & $-1.99$*** \\
                             & $0.25$ & $+14.5$***              & $-0.42$***              & $-0.8$\phantom{***}     & $-3.87$*** \\
\midrule
all tasks (60/60)            & $0.02$ & $+4.0$***               & $-0.13$***              & $+4.9$***               & $-0.07$\phantom{***} \\
                             & $0.05$ & $+6.4$***               & $-0.25$***              & $+9.0$***               & $-0.53$*** \\
                             & $0.10$ & $+6.8$***               & $-0.26$***              & $+5.8$*\phantom{**}     & $-1.00$*** \\
                             & $0.25$ & $+6.1$\phantom{***}     & $-0.45$***              & $-11.9$***              & $-3.27$*** \\
\bottomrule
\end{tabular}
\caption{\textbf{Coupling EMA axis}: median paired difference (EMA rate $\alpha$ $\!-\!$ frozen teacher, pp) at a fixed number of steps, pairs identical except for the coupling ($2$ rollout sources $\times$ $2$ KL directions $\times$ $3$ learning rates); Holm applied within each model, metric and rate; $n$ given as Qwen/Ministral.} 
\label{tab:app:stats-coupling}
\end{table}

\subsubsection{Teacher Coupling}\label{app:llm_results_ema}

Fig.~\ref{fig:axis2_appendix} extends Fig.~\ref{fig:llm:coupling}a with reverse KL in addition to forward KL (top), and with Ministral for both KL directions (bottom). All plots use teacher rollouts.

The per-task tests on matched pairs are reported in Tab.~\ref{tab:app:stats-coupling}.

\subsubsection{Token-Level Divergence}\label{app:llm_results_kl_dir}

Fig.~\ref{fig:llm:axis-tradeoffs:student:appendix} extends Fig.~\ref{fig:llm:axis-tradeoffs} with student instead of teacher rollouts.

The per-task tests on matched pairs are reported in Tab.~\ref{tab:app:stats-kl}.

\begingroup
\setlength{\intextsep}{4pt}
\begin{figure}[H]

    \centering
    \begin{subfigure}[b]{0.655\linewidth}
        \centering
        \includegraphics[scale=0.87]{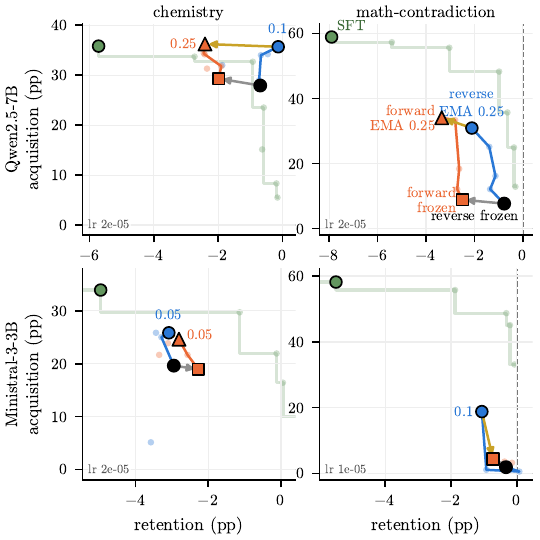}
        \caption{LLM experiments}
    \end{subfigure}\hfill
    \begin{subfigure}[b]{0.327\linewidth}
        \centering
        \includegraphics[scale=0.87]{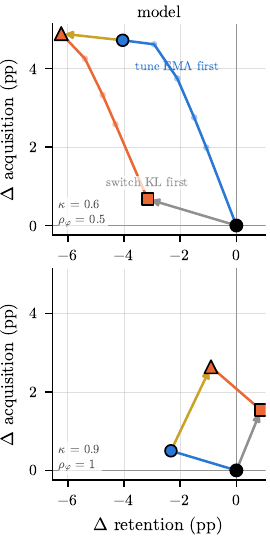}
        \caption{Controlled model}
    \end{subfigure}
    \vspace{-0.1cm}
    \caption{\textbf{Coupling and token-level divergence axes.}
(a)~Student rollouts, one learning rate per panel (corner note; as in Fig.~\ref{fig:llm:axis-tradeoffs}a). Routes from frozen reverse KL: the EMA coupling tuned under each direction (blue: reverse; orange: forward; best $\alpha$ labeled), the direction switch frozen (gray) and between the optima (gold). Green: the SFT grid and its Pareto front.
(b)~The model counterpart ($\lambda=2.5$, context strength and feature overlap in the corner notes), axes relative to frozen reverse KL.}
    \label{fig:llm:axis-tradeoffs:student:appendix}
\end{figure}

\begin{table}[H]
\centering
\small
\providecommand{\kv}[3]{\makebox[0pt][r]{$#1$}.\makebox[0pt][l]{$#2$#3}}
\providecommand{\seedv}[2]{{\scriptsize\textcolor{gray}{\makebox[4em][r]{(#1}{,}\makebox[4em][l]{\ #2)}}}}
{\setlength{\tabcolsep}{4pt}\renewcommand{\arraystretch}{0.96}
\begin{tabular}{l cc cc}
\toprule
& \multicolumn{2}{c}{Qwen2.5-7B} & \multicolumn{2}{c}{Ministral-3-3B} \\
\cmidrule(lr){2-3} \cmidrule(lr){4-5}
Task ($n$ pairs) & acquisition & retention & acquisition & retention \\
\midrule
tool-alpaca (30) & \kv{-0}{7}{} & \kv{-0}{9}{***} & \kv{-0}{8}{} & \kv{-0}{0}{} \\[-3.5pt]
 & \seedv{$-1.4$}{$-0.3$} & \seedv{$-0.8$***}{$-0.8$***} & \seedv{$-1.7$}{$-1.5$} & \seedv{$-0.1$}{$+0.1$} \\[0pt]
chemistry (30) & \kv{+1}{4}{**} & \kv{-0}{5}{***} & \kv{-0}{2}{} & \kv{+0}{3}{***} \\[-3.5pt]
 & \seedv{$+1.5$*}{$+1.2$**} & \seedv{$-0.6$***}{$-0.4$***} & \seedv{$-0.0$}{$+0.0$} & \seedv{$+0.4$***}{$+0.3$***} \\[0pt]
spatial (30) & \kv{+0}{3}{} & \kv{-0}{3}{***} & \kv{+6}{0}{*} & \kv{+0}{1}{} \\[-3.5pt]
 & \seedv{$-0.4$}{$+0.5$} & \seedv{$-0.4$***}{$-0.3$**} & \seedv{$+4.4$**}{$+5.4$} & \seedv{$+0.2$}{$+0.2$} \\[0pt]
spatial-contradiction (30) & \kv{-0}{4}{} & \kv{-0}{4}{***} & \kv{+0}{4}{} & \kv{+0}{2}{} \\[-3.5pt]
 & \seedv{$-0.5$}{$-0.2$} & \seedv{$-0.4$***}{$-0.4$***} & \seedv{$+1.2$}{$-0.0$} & \seedv{$+0.0$}{$+0.2$} \\[0pt]
math-contradiction (30) & \kv{+5}{0}{***} & \kv{-0}{9}{***} & \kv{+2}{6}{} & \kv{+0}{6}{**} \\[-3.5pt]
 & \seedv{$+5.5$***}{$+5.1$***} & \seedv{$-0.8$***}{$-0.8$***} & \seedv{$+0.9$}{$+2.6$} & \seedv{$+0.2$}{$+0.6$*} \\
\midrule
ordinary (90) & \kv{+0}{7}{} & \kv{-0}{5}{***} & \kv{+0}{3}{} & \kv{+0}{2}{***} \\[-3.5pt]
 & \seedv{$+0.2$}{$+0.6$} & \seedv{$-0.6$***}{$-0.5$***} & \seedv{$+0.6$}{$+0.0$} & \seedv{$+0.2$**}{$+0.3$***} \\[0pt]
contradictory (60) & \kv{+1}{1}{**} & \kv{-0}{5}{***} & \kv{+1}{0}{} & \kv{+0}{3}{**} \\[-3.5pt]
 & \seedv{$+1.0$**}{$+0.7$**} & \seedv{$-0.5$***}{$-0.5$***} & \seedv{$+1.1$*}{$+0.7$} & \seedv{$+0.1$*}{$+0.3$**} \\
\midrule
all tasks (150) & \kv{+0}{8}{**} & \kv{-0}{5}{***} & \kv{+0}{4}{} & \kv{+0}{2}{***} \\[-3.5pt]
 & \seedv{$+0.5$*}{$+0.6$**} & \seedv{$-0.5$***}{$-0.5$***} & \seedv{$+0.8$*}{$+0.2$} & \seedv{$+0.2$***}{$+0.3$***} \\
\midrule
all tasks, lr 5e-6 (50) & \kv{+0}{8}{} & \kv{-0}{3}{***} & \kv{-0}{2}{} & \kv{-0}{0}{} \\[-3.5pt]
 & \seedv{$+0.7$}{$+0.5$*} & \seedv{$-0.3$***}{$-0.3$***} & \seedv{$-0.4$}{$-0.9$} & \seedv{$-0.0$}{$+0.0$} \\[0pt]
all tasks, lr 1e-5 (50) & \kv{+0}{6}{} & \kv{-0}{5}{***} & \kv{+1}{1}{} & \kv{+0}{1}{*} \\[-3.5pt]
 & \seedv{$+0.1$}{$+0.5$} & \seedv{$-0.6$***}{$-0.5$***} & \seedv{$+1.3$}{$+1.1$} & \seedv{$+0.1$}{$+0.3$**} \\[0pt]
all tasks, lr 2e-5 (50) & \kv{+1}{3}{} & \kv{-1}{0}{***} & \kv{+1}{0}{*} & \kv{+1}{1}{***} \\[-3.5pt]
 & \seedv{$+1.5$}{$+1.2$} & \seedv{$-1.0$***}{$-0.9$***} & \seedv{$+1.2$*}{$+0.6$} & \seedv{$+0.9$***}{$+1.3$***} \\
\bottomrule
\end{tabular}
}
\caption{\textbf{Token-level divergence axis}: median paired difference (forward $\!-\!$ reverse, pp), pairs identical except for the KL direction ($2$ rollout sources $\times$ $5$ couplings $\times$ $3$ learning rates); the last three rows split the pooled pairs by learning rate. Below each value, in gray: the same median for each of the two seeds taken alone.}
\label{tab:app:stats-kl}
\end{table}
\endgroup

\subsubsection{Comparison with SFT}\label{app:llm_results_sft}

Tab.~\ref{tab:app:sft-crossing} compares self-distillation with SFT: each (rollout source, KL direction) family is shown at its highest acquisition among 5 runs (one per coupling rate, at the learning rate fixed per model in Sec.~\ref{sec:experimental_design}), against the SFT Pareto front over 7 runs (one per learning rate). Tab.~\ref{tab:app:sft-crossing-lr-free} repeats this comparison when the self-distillation run is also selected over the three learning rates (15 runs per family, against the same 7 for SFT).

\begin{table}[H]
\centering
\footnotesize
\setlength{\tabcolsep}{3pt}
\begin{tabular}{l r rrrr}
\toprule
Task & SFT$^\star$ & teacher, forward & teacher, reverse & student, forward & student, reverse \\
\midrule
\multicolumn{6}{l}{\emph{Qwen2.5-7B (lr 2e-5)}} \\
tool-alpaca           & $+25/\phantom{1}{-3.4}$ & \textcolor{tabblue}{{\boldmath$+28/{-2.3}$}} & \textcolor{tabblue}{$+24/{-0.7}$} & \textcolor{tabblue}{{\boldmath$+27/{-1.7}$}} & \textcolor{tabblue}{$+24/{-0.8}$} \\
chemistry             & $+36/\phantom{1}{-5.7}$ & \textcolor{tabblue}{$+33/{-0.8}$} & \textcolor{tabblue}{$+33/{-0.8}$} & \textcolor{tabblue}{{\boldmath$+36/{-2.4}$}} & \textcolor{tabblue}{$+36/{-0.1}$} \\
spatial               & $+72/\phantom{1}{-7.7}$ & \textcolor{tabblue}{$+61/{-1.6}$} & \textcolor{tabblue}{$+43/{-1.0}$} & \textcolor{tabblue}{$+54/{-0.6}$} & \textcolor{tabblue}{$+55/{-0.8}$} \\
spatial-contradiction & $+97/{-13.1}$ & \textcolor{tabblue}{$+88/{-1.8}$} & \textcolor{tabblue}{$+78/{-1.5}$} & $+26/{-0.8}$ & \textcolor{tabblue}{$+34/{-0.4}$} \\
math-contradiction    & $+59/\phantom{1}{-7.9}$ & \textcolor{tabblue}{$+51/{-2.8}$} & \textcolor{tabblue}{$+49/{-1.4}$} & $+34/{-3.3}$ & $+31/{-2.1}$ \\
\midrule
\multicolumn{6}{l}{\emph{Ministral-3-3B (lr 5e-6)}} \\
tool-alpaca           & $+62/\phantom{1}{-6.0}$ & \textcolor{tabblue}{$+42/{-0.3}$} & \textcolor{tabblue}{$+43/{-0.6}$} & \textcolor{tabblue}{$+25/{-0.1}$} & \textcolor{tabblue}{$+40/{-0.3}$} \\
chemistry             & $+34/\phantom{1}{-5.0}$ & $+18/{-0.4}$ & $+18/{-0.1}$ & \textcolor{tabblue}{$+18/{+0.6}$} & \textcolor{tabblue}{$+18/{+0.2}$} \\
spatial               & $+79/{-20.6}$ & \textcolor{tabblue}{$+42/{-0.8}$} & \textcolor{tabblue}{$+31/{-0.6}$} & \textcolor{tabblue}{$+32/{-0.0}$} & \textcolor{tabblue}{$+40/{+0.1}$} \\
spatial-contradiction & $+95/{-13.5}$ & \textcolor{tabblue}{$+32/{-0.4}$} & \textcolor{tabblue}{$+49/{-1.0}$} & \textcolor{tabblue}{$+17/{+0.1}$} & \textcolor{tabblue}{$+22/{+0.3}$} \\
math-contradiction    & $+58/\phantom{1}{-5.5}$ & $+35/{-1.3}$ & $+31/{-1.9}$ & $+12/{-0.6}$ & $+19/{-0.2}$ \\
\bottomrule
\end{tabular}
\caption{\textbf{Best self-distillation run against the SFT grid.} Each cell reports acquisition/retention (pp). For each model, task and (rollout source, KL direction) family, we show the run with the highest mean validation acquisition over the five coupling rates $\alpha$ (frozen included), at the per-model learning rate of Sec.~\ref{sec:experimental_design} (model header). SFT$^\star$ is likewise the SFT run with the highest mean validation acquisition over its learning-rate grid (Sec.~\ref{sec:experimental_design}), at the same number of epochs. Blue: beyond the SFT Pareto front, i.e., no SFT run is at least as good on both metrics; black: dominated by some SFT run. Bold: acquisition also at least as high as SFT$^\star$. }
\label{tab:app:sft-crossing}
\end{table}

\begin{table}[H]
\centering
\footnotesize
\setlength{\tabcolsep}{3pt}
\begin{tabular}{l r rrrr}
\toprule
Task & SFT$^\star$ & teacher, forward & teacher, reverse & student, forward & student, reverse \\
\midrule
\multicolumn{6}{l}{\emph{Qwen2.5-7B}} \\
tool-alpaca           & $+25/\phantom{1}{-3.4}$ & \textcolor{tabblue}{{\boldmath$+28/{-2.3}$}} & \textcolor{tabblue}{$+24/{-0.1}$} & \textcolor{tabblue}{{\boldmath$+27/{-1.7}$}} & \textcolor{tabblue}{$+25/{-0.1}$} \\
chemistry             & $+36/\phantom{1}{-5.7}$ & \textcolor{tabblue}{$+33/{-0.8}$} & \textcolor{tabblue}{$+33/{-0.8}$} & \textcolor{tabblue}{{\boldmath$+36/{-2.4}$}} & \textcolor{tabblue}{$+36/{-0.1}$} \\
spatial               & $+72/\phantom{1}{-7.7}$ & \textcolor{tabblue}{$+61/{-1.6}$} & \textcolor{tabblue}{$+43/{-1.0}$} & \textcolor{tabblue}{$+54/{-0.6}$} & \textcolor{tabblue}{$+55/{-0.8}$} \\
spatial-contradiction & $+97/{-13.1}$ & \textcolor{tabblue}{$+88/{-1.8}$} & \textcolor{tabblue}{$+78/{-1.5}$} & $+26/{-0.8}$ & \textcolor{tabblue}{$+34/{-0.4}$} \\
math-contradiction    & $+59/\phantom{1}{-7.9}$ & \textcolor{tabblue}{$+51/{-2.8}$} & \textcolor{tabblue}{$+49/{-1.4}$} & $+34/{-3.3}$ & $+31/{-2.1}$ \\
\midrule
\multicolumn{6}{l}{\emph{Ministral-3-3B}} \\
tool-alpaca           & $+62/\phantom{1}{-6.0}$ & \textcolor{tabblue}{$+42/{-0.3}$} & \textcolor{tabblue}{$+45/{-0.6}$} & \textcolor{tabblue}{$+40/{-0.6}$} & $+43/{-1.2}$ \\
chemistry             & $+34/\phantom{1}{-5.0}$ & \textcolor{tabblue}{$+31/{-3.5}$} & \textcolor{tabblue}{$+32/{-3.9}$} & $+25/{-2.8}$ & $+26/{-3.1}$ \\
spatial               & $+79/{-20.6}$ & $+56/{-9.2}$ & $+54/{-8.9}$ & \textcolor{tabblue}{$+51/{-1.0}$} & \textcolor{tabblue}{$+49/{-0.7}$} \\
spatial-contradiction & $+95/{-13.5}$ & \textcolor{tabblue}{$+76/{-2.8}$} & \textcolor{tabblue}{$+49/{-1.0}$} & \textcolor{tabblue}{$+27/{-0.7}$} & \textcolor{tabblue}{$+43/{-1.3}$} \\
math-contradiction    & $+58/\phantom{1}{-5.5}$ & $+40/{-4.5}$ & $+31/{-1.9}$ & $+12/{-0.6}$ & $+19/{-0.2}$ \\
\bottomrule
\end{tabular}
\caption{\textbf{Best self-distillation run against the SFT grid, with the learning rate also selected.} Same as Tab.~\ref{tab:app:sft-crossing}, except that each self-distillation cell is the run with the highest mean validation acquisition over the five coupling rates $\alpha$ and the three learning rates of Sec.~\ref{sec:experimental_design} ($15$ runs per family, against $7$ for SFT). The SFT grid and its best-acquiring run are unchanged; colors and bold as in Tab.~\ref{tab:app:sft-crossing}.}
\label{tab:app:sft-crossing-lr-free}
\end{table}

\subsection{Acquisition at the Final Checkpoint}\label{app:llm_results_same_ckpt}

In the paper, acquisition averages the last three saved checkpoints, while retention is evaluated at the final one (Sec.~\ref{sec:experimental_design}). As an ablation, we repeat the per-axis tests of Tabs.~\ref{tab:app:stats-rollout}, \ref{tab:app:stats-coupling} and~\ref{tab:app:stats-kl} with acquisition measured at the final checkpoint only. The trade-offs are unchanged: no cell changes value by more than 2~pp and no significant effect changes sign. Six cells change significance level (two of the 16 acquisition cells of Tab.~\ref{tab:app:stats-rollout}, one of the 64 of Tab.~\ref{tab:app:stats-coupling}, three of the 22 of Tab.~\ref{tab:app:stats-kl}). Tab.~\ref{tab:app:same-ckpt} lists them, together with the only sign change.

\begin{table}[!tp]
\centering
\small
\begin{tabular}{l l l rr}
\toprule
Axis & Model & Task & \multicolumn{1}{c}{mean of last three} & \multicolumn{1}{c}{final checkpoint} \\
\midrule
Rollout (Tab.~\ref{tab:app:stats-rollout}) & Ministral-3-3B & chemistry & $+2.0$**\phantom{*} & $+1.9$*\phantom{**} \\
Rollout (Tab.~\ref{tab:app:stats-rollout}) & Ministral-3-3B & ordinary & $+2.2$**\phantom{*} & $+2.0$*\phantom{**} \\
Coupling (Tab.~\ref{tab:app:stats-coupling}), $\alpha=0.02$ & Qwen2.5-7B & tool-alpaca & $+2.5$*\phantom{**} & $+2.1$\phantom{***} \\
KL direction (Tab.~\ref{tab:app:stats-kl}) & Qwen2.5-7B & chemistry & $+1.4$**\phantom{*} & $+0.8$\phantom{***} \\
KL direction (Tab.~\ref{tab:app:stats-kl}) & Qwen2.5-7B & all tasks & $+0.8$**\phantom{*} & $+0.7$*\phantom{**} \\
KL direction (Tab.~\ref{tab:app:stats-kl}) & Ministral-3-3B & all tasks & $+0.4$\phantom{***} & $+1.0$*\phantom{**} \\
KL direction (Tab.~\ref{tab:app:stats-kl}) & Ministral-3-3B & chemistry & $-0.2$\phantom{***} & $+0.2$\phantom{***} \\
\bottomrule
\end{tabular}
\caption{Acquisition at the final checkpoint: the acquisition cells of Tabs.~\ref{tab:app:stats-rollout}, \ref{tab:app:stats-coupling} and~\ref{tab:app:stats-kl} whose significance level \emph{or} sign changes when acquisition is measured at the final checkpoint instead of the mean over the last three; stars as in Tab.~\ref{tab:app:stats-rollout}. The two cells that lose significance are at $p = 0.06$; the sign change is not significant under either measure.}
\label{tab:app:same-ckpt}
\end{table}

\FloatBarrier
\subsection{Seed Robustness}\label{app:llm_results_seeds}

The main text averages every result over two seeds (13 and 31). App.~\ref{app:llm_results_third_seed} adds a third seed to Figs.~\ref{fig:llm:rollout}a and~\ref{fig:llm:coupling}a. The main text also measures acquisition on the validation set of each task, and the same set selects the configurations shown in some figures and tables (Figs.~\ref{fig:llm:rollout}a,~\ref{fig:llm:axis-tradeoffs}a; Tabs.~\ref{tab:app:sft-crossing},~\ref{tab:app:sft-crossing-lr-free}), such as the best coupling rate $\alpha^\star$ or the best SFT learning rate. App.~\ref{app:llm_results_test_split} trains the selected configurations of one panel of Fig.~\ref{fig:llm:axis-tradeoffs}a with two new seeds and evaluates them on an unused test split.

\subsubsection{Third Seed}\label{app:llm_results_third_seed}

\begin{figure}[!b]
    \centering
    \begin{subfigure}[t]{0.535\linewidth}
        \centering
        \includegraphics[width=\linewidth]{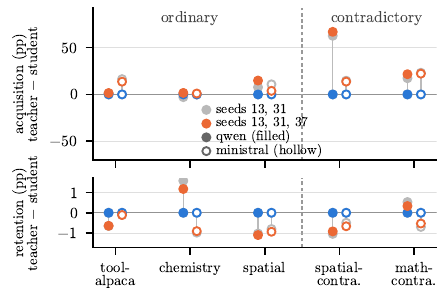}
        \caption{Rollout source, Fig.~\ref{fig:llm:rollout}a.}
    \end{subfigure}\hfill
    \begin{subfigure}[t]{0.453\linewidth}
        \centering
        \includegraphics[width=\linewidth]{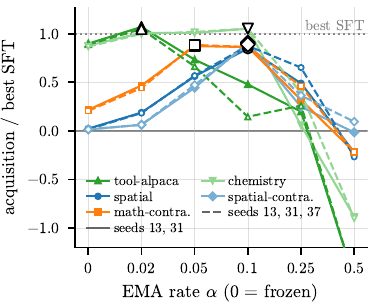}
        \caption{Coupling rate, Fig.~\ref{fig:llm:coupling}a.}
    \end{subfigure}
    \caption{\textbf{Figs.~\ref{fig:llm:rollout}a and~\ref{fig:llm:coupling}a with a third seed (37).} (a)~Gray: seeds 13 and 31, as in the main text. Colors: seeds 13, 31 and 37, at the same coupling rates. (b)~Solid: seeds 13 and 31. Dashed: seeds 13, 31 and 37, divided by the same best SFT gain. All other settings as in the main figures.}
    \label{fig:app:third-seed}

    \vspace{1.2em}
    \centering
    \includegraphics[width=\linewidth]{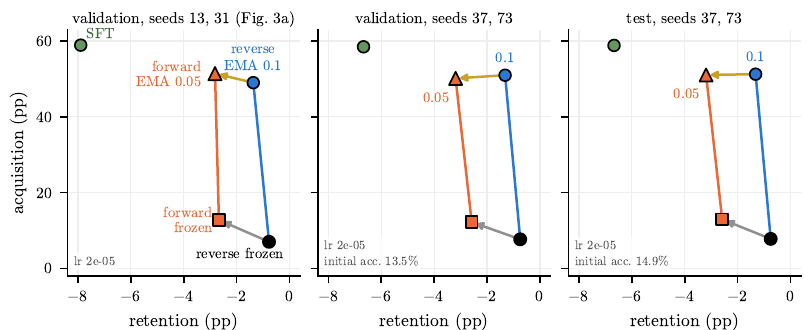}
    \caption{\textbf{Validation and test acquisition of the selected configurations on math-contradiction.} Math-contradiction panel of Fig.~\ref{fig:llm:axis-tradeoffs}a (Qwen, teacher rollouts, lr 2e-5), with the five configurations and their coupling rates fixed as selected on the validation set. Left: original runs (seeds 13, 31). Middle: two new seeds (37, 73), validation set. Right: the same runs as the middle panel, evaluated on the separate test split ($1{,}272$ problems); acquisition is measured from the initial test accuracy ($14.9\%$, against $13.5\%$ on validation), and retention, which does not depend on the task split, is unchanged.}
    \label{fig:app:val-vs-test}
\end{figure}

We train a third seed for every run of Figs.~\ref{fig:llm:rollout}a and~\ref{fig:llm:coupling}a. Fig.~\ref{fig:app:third-seed} compares the three-seed averages with the two-seed averages of the main text. On the rollout source (Fig.~\ref{fig:app:third-seed}a), teacher rollouts still win every pair of the contradictory tasks by a wide margin, and the retention differences barely change. On the coupling rate (Fig.~\ref{fig:app:third-seed}b), the best rate $\alpha^\star$ is unchanged on all five tasks, and every curve still rises and then falls. The two averages differ mostly past $\alpha^\star$, where the first two seeds already differ most.

\subsubsection{Selection and Evaluation on Separate Splits}\label{app:llm_results_test_split}

In the main text, the validation set both selects the configurations and measures their acquisition, which could overstate the reported acquisition. We check this on the math-contradiction panel of Fig.~\ref{fig:llm:axis-tradeoffs}a (Qwen, teacher rollouts, lr 2e-5).
This panel shows five configurations, with choices made on the validation set: reverse KL with a frozen teacher and at its best coupling rate ($\alpha^\star=0.10$), forward KL with a frozen teacher and at its best coupling rate ($\alpha^\star=0.05$), and SFT at its best learning rate (5e-5). We keep these choices, train each configuration again with two new seeds (37 and 73), and evaluate these new runs on the validation set and on an unused test split (App.~\ref{app:llm:math-construction}, Fig.~\ref{fig:app:val-vs-test}).

\textbf{Validation set against test split, new seeds (middle and right panels).} The two panels show the same runs, evaluated on each set. Test acquisition is $0.1$ to $0.9$ points higher than validation acquisition for the five configurations, and never lower. On this task, selecting on the validation set does not overstate acquisition.

\textbf{Original seeds against new seeds, validation set (left and middle panels).} The new seeds reproduce the two large effects of the main text. Tuning the coupling under reverse KL adds $+43.3$ acquisition for a retention loss of $0.6$ ($+42.0$ for $0.6$ on the original seeds), and switching to forward KL with a frozen teacher adds $+4.5$ for $1.8$ ($+5.8$ for $1.9$). The smallest effect changes sign: between the two tuned couplings, forward KL gains $+2.4$ on the original seeds and loses $0.8$ on the new ones ($0.2$ on test). These values are below the difference between the two seeds of a single configuration (up to $2.5$ points).

The conclusions of Sec.~\ref{sec:llm_results_kl_dir} on Qwen therefore hold with new seeds and on unseen examples.

\FloatBarrier
\subsection{Other Privileged Contexts}\label{app:llm_context_ablations}

In the main grid, the privileged context is the full reference response (Sec.~\ref{sec:taxonomy}). To test whether our findings depend on this choice, we repeat the coupling and KL-direction comparisons of Figs.~\ref{fig:llm:coupling}a and~\ref{fig:llm:axis-tradeoffs}a with two other contexts: the final answer alone (App.~\ref{app:llm_context_hint}) and feedback on the student's response (App.~\ref{app:llm_context_feedback}).

\subsubsection{Final Answer as Privileged Context}\label{app:llm_context_hint}
\begin{figure}[!htb]
    \centering
    \begin{subfigure}[t]{0.376\linewidth}
        \centering
        \includegraphics[width=\linewidth]{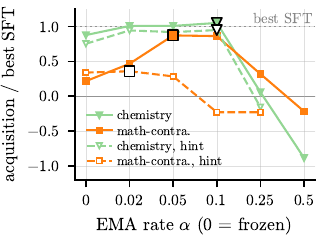}
        \caption{Coupling panel of Fig.~\ref{fig:llm:coupling}a: full reference (solid) and hint (dashed).}
    \end{subfigure}\hfill
    \begin{subfigure}[t]{0.607\linewidth}
        \centering
        \includegraphics[width=\linewidth]{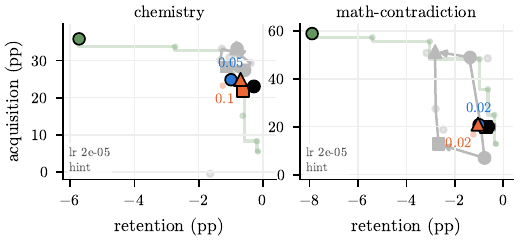}
        \caption{KL panels of Fig.~\ref{fig:llm:axis-tradeoffs}a with the hint; main-text routes in gray.}
    \end{subfigure}
    \caption{\textbf{With a hint as context, the findings hold on chemistry, but coupling no longer raises acquisition on math-contradiction.} Chemistry and math-contradiction, Qwen, teacher rollouts, lr 2e-5; points average two seeds.}
    \label{fig:app:final-answer}
\end{figure}

We first test how self-distillation behaves when we give 
the teacher only a hint: the final answer (the answer letter on chemistry, the base-9 result on math-contradiction). To do so, we repeat our experiments on Qwen, with teacher rollouts, lr 2e-5, the same training horizons and the five coupling rates (Fig.~\ref{fig:app:final-answer}). The hint is given as follows:

\begin{tcolorbox}[width=\textwidth,colframe=black!75!white,title=Teacher user message with the hint (chemistry and math-contradiction).]
\begin{Verbatim}[fontsize=\scriptsize,baselinestretch=0.9,breaklines,breakanywhere]
<TASK INPUT>

The correct final answer to this problem is: <FINAL ANSWER>

Now answer the original question with a response of your own, including the thinking process, and keep the requested final-answer format.
\end{Verbatim}
\end{tcolorbox}

On chemistry, both comparisons give the same results as in the main text. The coupling curve keeps its shape: a frozen teacher is already close to the best rate, and the fastest rate falls below the untrained accuracy (Fig.~\ref{fig:app:final-answer}a). 
With a frozen teacher and at the best coupling rate, switching to forward KL changes acquisition by at most about one point and retention by less than half a point (Fig.~\ref{fig:app:final-answer}b).
With less information in the context, acquisition peaks lower than with the full reference response.

On math-contradiction, the coupling curve is now flat as well: coupling no longer helps (Fig.~\ref{fig:app:final-answer}a). The hint contains no intermediate calculations, so the teacher has to write them itself. 
Tab.~\ref{tab:app:base9-share} reports whether the teacher and the student write these calculations in base 9, as the task requires, or in base 10. With the full reference, the share of the final student's calculations written in base 9 rises with the coupling rate, from 7\% with a frozen teacher to 89\% at $\alpha=0.05$.
With the hint, it stays below 4\%: neither the teacher nor the student writes calculations in base 9, at any coupling rate.
The gain from coupling that the main text reports on math-contradiction therefore comes with calculations written in base 9, which appear with the full reference and do not appear with the hint.

\begin{table}[!h]
\centering
\footnotesize
\caption{\textbf{Share of calculations written in base 9 on math-contradiction} (\%). In every response, we collect all the written steps of the form $a+b=c$, $a-b=c$ or $a\times b=c$ between integers, and check each one in base 9 and in base 10. Steps that are true in both bases (e.g. $1+3=4$) or in neither do not tell the base and are left out. The table gives, over all responses, the share of the remaining steps that are true in base 9. For example, $12-4=7$ is true in base 9 only and $12-4=8$ in base 10 only. Study with: Qwen, teacher rollouts, forward KL, lr 2e-5. The teacher column uses the initial model; student columns use the final checkpoint, averaged over two seeds, on the $500$ validation examples.}%
\label{tab:app:base9-share}
\begin{tabular}{l c ccccc}
\toprule
 & & \multicolumn{5}{c}{student, by coupling rate $\alpha$} \\
\cmidrule(lr){3-7}
Privileged context & initial teacher & frozen & $0.02$ & $0.05$ & $0.10$ & $0.25$ \\
\midrule
full reference & $18$ & $7$ & $20$ & $89$ & $97$ & $100$ \\
hint & $0$ & $0$ & $1$ & $0$ & $3$ & $3$ \\
\bottomrule
\end{tabular}
\end{table}

\vspace{3cm}
\subsubsection{Feedback as Privileged Context}\label{app:llm_context_feedback}

We then give the teacher feedback on the student's response. The context is then written during training: a feedback writer $f$ reads the input, the student's response $\bar y$ and optionally the reference response, and returns a short text that the teacher of Eq.~\ref{eq:distributions} reads as context,
\begin{equation}
c=f\big(x,\bar y,y^{\mathrm{ref}}\big).
\label{eq:feedback}
\end{equation}
The writer can be an environment, which returns for instance runtime errors or failing unit tests~\citep{hubotter2026sdpo}, or a model, which critiques or summarizes the student's responses~\citep{song2026rltf,wang2026skillsd,ye2026opcd}. We use two writers: an external model, GPT-5 mini (\emph{external feedback}), and the student itself, with its current weights (\emph{self-feedback}), which then changes during training along with the student. Both are asked for a few short instructions that point out the main error and how to correct it, without giving a full solution or the final answer (templates below). Since $c$ depends on $\bar y$, it changes as the student's errors change. The controlled model of Sec.~\ref{sec:toy_model} does not represent this case, since the model shifts the teacher toward the new task by an amount fixed before training.

Feedback is written on a student response, so we train with student rollouts: the teacher takes the feedback as privileged context and corrects the student on the response that this feedback describes.
Teacher rollouts would also be possible, as in RLTF~\citep{song2026rltf}, but they are less direct: the teacher writes a new response after reading the feedback, so each example needs one more generation, and the student is trained on a response that the feedback does not describe. As in App.~\ref{app:llm_context_hint}, we use Qwen, lr 2e-5 and the coupling rates of the main grid, with each task's standard number of epochs (Sec.~\ref{sec:experimental_design}). Self-feedback is run on chemistry and math-contradiction, and external feedback on chemistry only: on math-contradiction, preliminary runs gained little accuracy, and GPT-5 mini struggled to accept the base-9 convention of the reference, which made its reasoning much longer and the runs too costly.

\begin{tcolorbox}[splitbox, width=\textwidth,colframe=black!75!white,title=External feedback: request to GPT-5 mini (top) and teacher user message (bottom).]
\begin{Verbatim}[fontsize=\scriptsize,baselinestretch=0.9,breaklines,breakanywhere]
You provide corrective guidance for another attempt at a task.

Read the task, attempted response, and reference below. The reference specifies the intended target for this training example. Respect its task-specific conventions even when they differ from usual conventions.

Write one to three short numbered instructions, at most 120 words in total.

Requirements:
- Address a specific substantive error or omission in the attempt.
- Identify the relevant reasoning step, constraint, or output requirement and explain what should change.
- Preserve valid reasoning; do not demand a different approach merely because it differs from the reference.
- Make each instruction understandable from the task alone, without access to the attempted response or reference.
- Do not provide a complete replacement solution, reproduce the reference, or state the final answer directly.
- If the attempt is correct, briefly identify what should be preserved; do not invent an error.
- Output only the numbered instructions.

TASK:
<TASK INPUT>

ATTEMPTED RESPONSE:
<STUDENT RESPONSE>

REFERENCE:
<REFERENCE RESPONSE>
\end{Verbatim}
\tcblower
\begin{Verbatim}[fontsize=\scriptsize,baselinestretch=0.9,breaklines,breakanywhere]
<TASK INPUT>

Guidance for solving this task:
<FEEDBACK>

Solve the task, using the guidance where relevant and following the required output format.
\end{Verbatim}
\end{tcolorbox}

\begin{tcolorbox}[width=\textwidth,colframe=black!75!white,title=Self-feedback: request to the student (top) and teacher user message (bottom).]
\begin{Verbatim}[fontsize=\scriptsize,baselinestretch=0.9,breaklines,breakanywhere]
Compare the attempt with the reference and write 1-2 useful feedback bullets (at most 80 words).
Follow the reference's task conventions, even if unusual. Identify the main error and a specific correction. Preserve valid reasoning. If the attempt is correct, say what to preserve.
Make the feedback understandable from the task alone. Do not refer to the unseen attempt or reference. Do not write a full solution or the final answer.
Output only the feedback.

Task:
<TASK INPUT>

Attempt:
<STUDENT RESPONSE>

Reference:
<REFERENCE RESPONSE>
\end{Verbatim}
\tcblower
\begin{Verbatim}[fontsize=\scriptsize,baselinestretch=0.9,breaklines,breakanywhere]
<TASK INPUT>

Feedback for solving this task:
<FEEDBACK>

Answer the task using this feedback and the required output format.
\end{Verbatim}
\end{tcolorbox}

\begin{figure}[!b]
    \centering
    \vspace{-0.6cm}
    \begin{subfigure}[t]{\linewidth}
        \centering
        \includegraphics[width=0.89\linewidth, trim=0 0 43 0, clip]{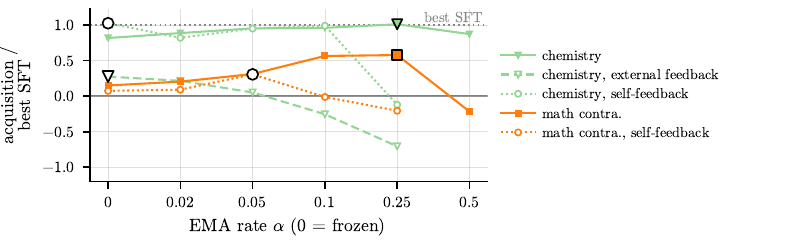}
        \caption{Coupling, forward KL: full reference (solid), external feedback (dashed) and self-feedback (dotted).}
    \end{subfigure}

    \begin{subfigure}[t]{\linewidth}
        \centering
        \includegraphics[width=\linewidth]{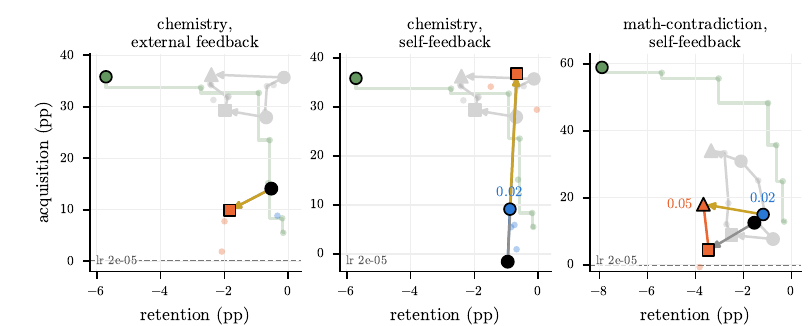}
        \caption{KL panels of Fig.~\ref{fig:llm:axis-tradeoffs}a; full-reference routes in gray. Rates below the untrained accuracy lie outside the frame.}
    \end{subfigure}
    \caption{\textbf{With feedback as context, acquisition peaks at a slow coupling rate, and forward KL keeps its retention cost except with self-feedback on chemistry.} Qwen, student rollouts, lr 2e-5, standard epochs (2 for chemistry, 4 for math-contradiction); points average two seeds. Unlike Figs.~\ref{fig:llm:coupling}a and~\ref{fig:llm:axis-tradeoffs}a, all curves, including the full reference, use student rollouts, and (a) uses the standard epochs rather than the same number of steps for every task: each feedback curve is to be compared with the full-reference curve of its own task, not across tasks. Unlike the full reference, the feedback text that the teacher reads changes during training with the student's responses, so we do not expect the same acquisition pattern: acquisition peaks at slower rates, and the faster rates fall below the untrained accuracy (a).}
    \label{fig:app:feedback}
\end{figure}

On chemistry, a frozen teacher is the best rate with both kinds of feedback, and the fastest rate falls below the untrained accuracy (Fig.~\ref{fig:app:feedback}a). With the full reference under student rollouts, the curve stays nearly flat up to $\alpha=0.25$.
With external feedback, switching to forward KL costs retention, as on Qwen in the main text (Sec.~\ref{sec:llm_results_kl_dir}), and here also lowers acquisition (Fig.~\ref{fig:app:feedback}b). With self-feedback, at the best rates, it costs no retention and raises acquisition up to the best SFT gain.

On math-contradiction, self-feedback also moves the peak to a slower rate, $\alpha=0.05$ (against $\alpha=0.25$ with the full reference; Fig.~\ref{fig:app:feedback}a). The Qwen result of Sec.~\ref{sec:llm_results_kl_dir} holds: switching to forward KL costs about two points of retention at every coupling rate, and adds a few points of acquisition at the best rates (Fig.~\ref{fig:app:feedback}b).

\subsection{Third Model Family}\label{app:llm_results_gemma}

\begin{figure}[!htb]
    \centering
    \includegraphics[scale=1]{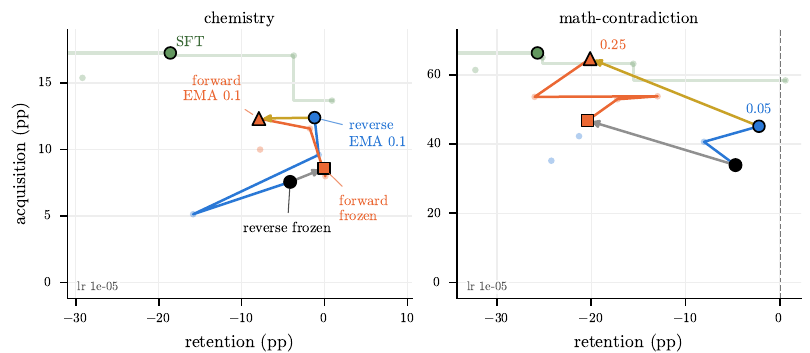}
    \caption{\textbf{On math-contradiction, forward KL costs retention on Gemma, as on Qwen; on chemistry, the differences are within the variation across seeds.} Panels of Fig.~\ref{fig:llm:axis-tradeoffs}a with Gemma-4-12B: teacher rollouts, lr 1e-5, standard epochs (2 for chemistry, 4 for math-contradiction); points average two seeds. Math-contradiction uses the template above. Rates far below the untrained accuracy lie outside the frame.}
    \label{fig:app:gemma-fig3a}
\end{figure}

The main grid uses two models, which differ in the retention cost of forward KL and thus in the axis to tune first: on Qwen, tuning the coupling under reverse KL costs little retention, whereas on Ministral switching to forward KL costs none (Sec.~\ref{sec:llm_results_kl_dir}). To see on which side a third model family falls, we repeat the comparison of Fig.~\ref{fig:llm:axis-tradeoffs}a with Gemma-4-12B (Gemma), teacher rollouts, lr 1e-5, two seeds, and the training horizons and coupling rates of the main grid (Fig.~\ref{fig:app:gemma-fig3a}).

With the teacher template of the main grid (App.~\ref{prompts:teacher}), Gemma learned little on math-contradiction (at most $16$ points of acquisition): its teacher often questioned the base-9 arithmetic of the reference and frequently looped until the maximum response length. We therefore add two sentences to the template, stating that the reference is correct and should not be corrected:

\begin{tcolorbox}[width=\textwidth,colframe=black!75!white,title=Teacher user message for Gemma on math-contradiction (third paragraph added).]
\begin{Verbatim}[fontsize=\scriptsize,baselinestretch=0.9,breaklines,breakanywhere]
<TASK INPUT>

Here is a verified reference solution for the same problem:
<REFERENCE RESPONSE>

The reference solution is correct. Its arithmetic is authoritative even where it looks unusual; do not try to correct it.

Now solve the original problem yourself. Keep the requested final-answer format.
\end{Verbatim}
\end{tcolorbox}

Gemma loses much more retention than Qwen and Ministral, and varies much more across seeds (Fig.~\ref{fig:app:gemma-fig3a}): the two seeds of one configuration differ by a median of $5$ points (against $0.2$ for Qwen and $0.7$ for Ministral on the panels of Fig.~\ref{fig:llm:axis-tradeoffs}a), mostly through IFEval and MMLU.

On math-contradiction, Gemma behaves like Qwen. Tuning the coupling under reverse KL adds about ten points of acquisition without losing retention. Switching to forward KL adds at least as much acquisition, but costs more than fifteen points of retention, both with a frozen teacher and at the best rates, where this cost holds for each seed. The coupling is thus the axis to tune first.

On chemistry, tuning the coupling raises acquisition for both KL directions and both seeds, by about four points on average. 
The two directions reach the same acquisition within one point. 
Forward KL retains four points more than reverse KL with a frozen teacher, and seven points fewer at the best rates. Both differences are within the variation across seeds (up to $16$ points on these configurations), so two seeds are too few to tell which axis to tune first on this task.

\section{Experimental Setting for the Controlled Model}\label{toy_experimental_settings}

This appendix specifies the controlled model of Sec.~\ref{sec:toy_model}: its fixed features and shared readout (App.~\ref{app:toy:backbone}), target and contextual teacher (App.~\ref{app:toy:construction}), training objective (App.~\ref{app:toy:objective}), and evaluation protocol (App.~\ref{app:toy:protocol}). The supporting experiments in App.~\ref{app:controlled_ablations} test the rollout-source, teacher-coupling, and KL-direction regimes discussed in the main text. The task controls $\lambda$, $\kappa$, and $\rho_\phi$ are varied separately from the three self-distillation axes.
\subsection{Finite autoregressive backbone and task geometry}
\label{app:toy:backbone}
\label{app:toy:task-geometry}

A response $\bar y=(\bar y_1,\ldots,\bar y_T)$ contains $T$ tokens from $\{1,\ldots,K\}$. Its nonterminal prefix states are
\[
\mathcal S=\bigcup_{t=1}^{T}\{1,\ldots,K\}^{t-1},\qquad s=\bar y_{<t}.
\]
The empty prefix is the root, and appending token $a$ moves from $s$ to $s\Vert a$. For a fixed feature map $\phi:\mathcal S\rightarrow\mathbb R^D$ and readout $W\in\mathbb R^{K\times D}$, the induced autoregressive policy is 
\[
\pi_W^\phi(\cdot\mid s)=\operatorname{softmax}(W\phi(s)).
\]
In particular, the old-task features are sampled independently as
$\phi_{\rm old}(s)=\sqrt{D}\,g_s/\|g_s\|_2$ with
$g_s\sim\mathcal N(0,I_D)$, giving
$\|\phi_{\rm old}(s)\|_2=\sqrt{D}$.

For random task seed $j$, let $(\widetilde W_0)_{a\ell}\sim\mathcal N(0,1/D)$ independently and define
\[
C_K=I_K-\frac1K\mathbf1\mathbf1^\top,\qquad
W_0=\lambda C_K\widetilde W_0.
\]
Centering removes the action-independent logit direction, while $\lambda>0$ sharpens the initial policy without changing token rankings. We therefore call $\lambda$ the \emph{initial-policy concentration}: it controls how strongly established the initial token preferences are, not the optimizer step size. Student and teacher start from the same readout, $W_0^\star=W_0$, and the behavior to retain is
\[
r_{\rm old}(\cdot\mid s)=\pi_{W_0}^{\phi_{\rm old}}(\cdot\mid s).
\]

The feature-overlap coefficient $\rho_\phi$ controls representation reuse.
For each state, independently draw $h_s\sim\mathcal N(0,I_D)$, remove its
component along the old-task feature, and normalize the residual:
\[
u_s=h_s-\frac{h_s^\top\phi_{\rm old}(s)}
{\|\phi_{\rm old}(s)\|_2^2}\phi_{\rm old}(s),
\qquad
\phi_\perp(s)=\sqrt D\,\frac{u_s}{\|u_s\|_2}.
\]
We then define the new-task feature map statewise as
\begin{equation}\label{eq:feat_overlap}
\phi_{\rm new}(s)=\rho_\phi\phi_{\rm old}(s)
+\sqrt{1-\rho_\phi^2}\,\phi_\perp(s).
\end{equation}

Since $\phi_\perp(s)\perp\phi_{\rm old}(s)$ and both have norm $\sqrt D$,
$\|\phi_{\rm new}(s)\|_2=\sqrt D$ and
$\cos(\phi_{\rm new}(s),\phi_{\rm old}(s))=\rho_\phi$ exactly.

The reference setting uses $\rho_\phi=0.5$. This is a statewise mixing control and should not be interpreted as a guarantee that the global spans of the two feature sets are orthogonal when $\rho_\phi=0$.

Both $\phi_{\rm old}$ and $\phi_{\rm new}$ are fixed throughout training,
while the two tasks share the trainable readout $W$.
Adaptation can therefore interfere with retained behavior even though
the feature maps themselves do not learn.

A separate readout-compatibility control $\rho_W\in[-1,1]$ constructs an
auxiliary target-generating readout. To define its orthogonal component,
independently draw
$(\widetilde W_\perp)_{a\ell}\sim\mathcal N(0,1/D)$ and center it as
$G_\perp=C_K\widetilde W_\perp$. We then remove its component along $W_0$
in Frobenius geometry and rescale the residual:
\[
U_\perp
=
G_\perp-
\frac{\langle G_\perp,W_0\rangle_F}{\|W_0\|_F^2}W_0,
\qquad
W_\perp=
\frac{\|W_0\|_F}{\|U_\perp\|_F}U_\perp,
\]
where $\langle A,B\rangle_F=\operatorname{tr}(A^\top B)$. The auxiliary readout is then
\[
W_{\rm task}(\rho_W)
=
\rho_W W_0+\sqrt{1-\rho_W^2}\,W_\perp.
\]
By construction,
$\langle W_\perp,W_0\rangle_F=0$ and
$\|W_\perp\|_F=\|W_0\|_F$, so
\[
\frac{\langle W_{\rm task}(\rho_W),W_0\rangle_F}
{\|W_{\rm task}(\rho_W)\|_F\|W_0\|_F}
=\rho_W.
\]
The reference construction sets $\rho_W=0$;
Appendix~\ref{rho_W_ablation} uses nonzero values as a controlled construction verification. 

Thus $\rho_\phi$ controls feature reuse, whereas $\rho_W$ controls compatibility of the mapping used to create the adaptation target.

\subsection{Attainable target and contextual teacher}
\label{app:toy:construction}
\label{app:toy:context-strength}

Using this auxiliary readout, at each prefix the two highest-scoring tokens under
$W_{\rm task}\phi_{\rm new}(s)$ receive prescribed masses $m_1=0.30$ and $m_2=0.15$, with the remaining mass spread uniformly over the other $K-2$ tokens. Denote this positive soft distribution by $\widetilde r(\cdot\mid s)$. Because an arbitrary collection of statewise distributions need not be representable by one shared readout, we project centered log probabilities into the readout family:
\begin{equation}
W_{\rm ref}\in\arg\min_W\sum_{s\in\mathcal S}
\left\|W\phi_{\rm new}(s)-C_K\log\widetilde r(\cdot\mid s)\right\|_2^2,
\qquad
r_{\rm new}(\cdot\mid s)=\pi_{W_{\rm ref}}^{\phi_{\rm new}}(\cdot\mid s).
\label{eq:toy:target-projection}
\end{equation}
The final reference $r_{\rm new}$, not the unprojected prescription, defines the adaptation task. It is therefore attainable by construction, while simultaneous old-task preservation is not guaranteed. 

Before adaptation, the context-free student applies the initial readout
$W_0$ to the new-task features:
\[
p_0(\cdot\mid s)=\pi_{W_0}^{\phi_{\rm new}}(\cdot\mid s).
\]

The \emph{context strength} $\kappa\in[0,1]$ specifies the fraction of the initial discrepancy from $r_{\rm new}$ removed by privileged information. Define
\begin{equation}
\widehat q_{\kappa,0}(\cdot\mid s)
=(1-\kappa)p_0(\cdot\mid s)+\kappa r_{\rm new}(\cdot\mid s).
\label{eq:toy:initial-context}
\end{equation}
For total variation $\operatorname{TV}(u,v)=\tfrac12\sum_a|u_a-v_a|$,
\[
\operatorname{TV}(\widehat q_{\kappa,0},r_{\rm new})
=(1-\kappa)\operatorname{TV}(p_0,r_{\rm new}),
\]
so $\kappa$ has an exact initial behavioral interpretation in the clean construction.

To realize this correction through privileged input rather than a fixed
probability mixture, let $w_a^\top$ denote row $a$ of $W$, take token $K$
as a logit reference, and define
\[
R(W)=\begin{bmatrix}(w_1-w_K)^\top\\ \vdots\\(w_{K-1}-w_K)^\top\end{bmatrix},
\qquad
[\delta_\kappa(s)]_a=
\log\frac{\widehat q_{\kappa,0}(a\mid s)}{\widehat q_{\kappa,0}(K\mid s)}
-\log\frac{p_0(a\mid s)}{p_0(K\mid s)}.
\]
With $R_0=R(W_0)$, the minimum-norm teacher-only feature displacement is
\begin{equation}
\psi_\kappa(s)=R_0^+\delta_\kappa(s),
\label{eq:toy:context-calibration}
\end{equation}
where $R_0^+$ is the Moore--Penrose pseudoinverse. In the reference dimensions $D\ge K-1$ and the random initialization gives full row rank almost surely, so the initial teacher is calibrated exactly. Thereafter the fixed displacement is processed by the evolving teacher readout:
\begin{equation}
\begin{aligned}
p_i(\cdot\mid s)&=\operatorname{softmax}(W_i\phi_{\rm new}(s)),\\
q_i(\cdot\mid s)&=\operatorname{softmax}\!\left(W_i^\star[\phi_{\rm new}(s)+\psi_\kappa(s)]\right).
\end{aligned}
\label{eq:toy:policies}
\end{equation}
Thus Eq.~\ref{eq:toy:initial-context} calibrates only $i=0$; it does not replace the teacher by a fresh probability mixture after every update.

\subsection{Exact prefix weighting and self-distillation axes}
\label{app:toy:objective}

For an autoregressive policy $\pi$, let $\nu_\pi(s)$ be the probability of reaching prefix $s$:
\[
\nu_\pi(\varnothing)=1,\qquad
\nu_\pi(s\Vert a)=\nu_\pi(s)\pi(a\mid s).
\]
Nonterminal occupancies sum to $T$. 

For pure student or teacher rollouts, let $\mu_i=p_i$ or $q_i$, respectively, and use the corresponding exact prefix occupancy $\nu_{\mu_i}$. These are the two rollout-source settings studied in the paper.

With all response tokens supervised, the taxonomy objective becomes
\begin{equation}
\mathcal L_i(W_i)=\frac1T\sum_{s\in\mathcal S}
\operatorname{sg}[\nu_{\mu_i}(s)]\,
 d\!\left(\operatorname{sg}[q_i(\cdot\mid s)],p_i(\cdot\mid s)\right),
\label{eq:toy:objective}
\end{equation}
where $d$ is forward or reverse KL and $\operatorname{sg}$ denotes stop-gradient. Only $W_i$ is optimized. Prefix occupancies and teacher predictions are recomputed after the student/teacher update, while Eq.~\ref{eq:ema} controls teacher coupling independently of rollout source and KL direction. Using exact prefix occupancies removes the trajectory-sampling noise present in sampled rollouts.

\subsection{Reference protocol, selection, and metrics}
\label{app:toy:protocol}
\label{app:toy:metrics}

The reference setting is summarized in Tab.~\ref{tab:mechanistic_reference}. Each reported random task seed regenerates the controlled task construction; method comparisons are paired on that seed.

\begin{table}[t]
\centering\small
\caption{Reference configuration of the controlled model.}
\label{tab:mechanistic_reference}
\begin{tabular}{@{}lll@{}}\toprule
Quantity & Reference & Role\\\midrule
$\lambda$ & $2.5$ & Initial-policy concentration\\
$\kappa$ & $0.6$ & Context strength \\
$\rho_\phi$ & $0.5$ & Old/new feature overlap\\
$\rho_W$ & $0$ & Auxiliary readout compatibility\\
$(T,K,D)$ & $(4,8,64)$ & Depth, branching factor, feature dimension\\
$(m_1,m_2)$ & $(0.30,0.15)$ & Pre-projection target masses\\
Optimizer & Adam, $10^{-3}$ & Student-readout optimization\\
Training & $200$ updates & Reference horizon\\
Weight decay / clipping & $0$ / $10$ & Optimizer controls\\
Occupancy & Exact & Prefix-visitation probabilities\\\bottomrule
\end{tabular}
\end{table}

Coupling selection is stated locally for each experiment in App.~\ref{app:controlled_ablations}. For cross-fitted comparisons, task seeds are divided into two folds using $\operatorname{fold}(j)=j\bmod 2$; the rate maximizing mean acquisition on one fold is evaluated on the other. Descriptive grid optima instead maximize sample-mean acquisition over all seeds and are identified as such. Fixed-coupling contrasts do not select $\alpha$. The main factorial completion and the dedicated rollout/coupling sweeps use $96$ task seeds, while targeted controls report their own cohort sizes. Cohorts are paired within experiments and are not pooled across experiments.

Define $\operatorname{Sim}(u,v)=1-\operatorname{TV}(u,v)$. New-task similarity ($S_{\rm new}$) and deployment-weighted acquisition ($\Delta_{\rm new}$) are
\[
S_{\rm new}(i)=\frac1T\sum_s\nu_{p_i}(s)
\operatorname{Sim}(p_i(\cdot\mid s),r_{\rm new}(\cdot\mid s)),
\qquad
\Delta_{\rm new}=S_{\rm new}(i_{\rm end})-S_{\rm new}(0).
\]
A complementary diagnostic holds the weighting distribution fixed to the target policy:
\[
S_{\rm new}^{\rm tgt}(i)=\frac1T\sum_s\nu_{r_{\rm new}}(s)
\operatorname{Sim}(p_i(\cdot\mid s),r_{\rm new}(\cdot\mid s)),
\qquad
\Delta_{\rm new}^{\rm tgt}=S_{\rm new}^{\rm tgt}(i_{\rm end})-S_{\rm new}^{\rm tgt}(0).
\]
We call $\Delta_{\rm new}^{\rm tgt}$ \emph{target-path fidelity}: it measures the same local distributional match as $\Delta_{\rm new}$, but weights prefixes by the fixed occupancy induced by the target policy rather than by the deployed student.

Old-task similarity uses the fixed old-reference occupancy,
\[
S_{\rm old}(i)=\frac1T\sum_s\nu_{r_{\rm old}}(s)
\operatorname{Sim}(\pi_{W_i}^{\phi_{\rm old}}(\cdot\mid s),r_{\rm old}(\cdot\mid s)),
\qquad
\Delta_{\rm old}=S_{\rm old}(i_{\rm end})-1.
\]
Thus larger $\Delta_{\rm old}$ means better retention, with forgetting being defined as: $F_{\rm old}=-\Delta_{\rm old}$. These are controlled-model distributional quantities, playing a similar role to the broad benchmark average for our LLM experiments.

Contextual-teacher utility on student-visited states is
\[
U_{\rm ctx}(i)=\frac1T\sum_s\nu_{p_i}(s)
\operatorname{Sim}(q_i(\cdot\mid s),r_{\rm new}(\cdot\mid s)).
\]
Similarity changes and forgetting are stored as fraction-scale quantities; figures that label such differences in \emph{points} multiply them by $100$. Absolute teacher utility remains on its native $[0,1]$ scale. Unless otherwise stated, method contrasts are formed within task seed before aggregation, and reported 95\% confidence intervals use the normal approximation $\bar x\pm1.96\,\mathrm{SE}$ over the resulting task-seed values.

Checks of exact versus sampled prefix occupancy and of the reference $(T,K,D)$ choices are reported in App.~\ref{app:robustness}.

\section{Controlled-Model Results and Regime Tests}
\label{app:controlled_ablations}

This appendix contains the controlled-model experiments that support the three result axes of Sec.~\ref{sec:llm_results}: rollout source (App.~\ref{app:rollout_controls}), teacher coupling (App.~\ref{app:coupling_controls}), and token-level divergence (App.~\ref{app:kl_controls}). We then report scoped numerical and structural checks in App.~\ref{app:robustness}. The model construction and metrics are defined in App.~\ref{toy_experimental_settings}. Unless stated otherwise, experiments use the reference task and optimization settings in Tab.~\ref{tab:mechanistic_reference}. The 18-rate coupling sweeps use $\alpha\in\{0\}\cup\{0.000625\,2^{j/2}:j=0,\ldots,16\}$; controls using smaller grids state them explicitly. Acquisition and retention changes are reported in points, i.e. the corresponding similarity changes multiplied by $100$; teacher utility remains on its native $[0,1]$ scale and entropy is reported in nats. Confidence intervals are pointwise normal 95\% intervals over task seeds; method-difference intervals use within-seed paired contrasts. They are not corrected for multiple comparisons. Results from distinct seed cohorts are not pooled; each experiment states its cohort and coupling-selection rule locally.

\subsection{Rollout source: deployment transfer and teacher utility}
\label{app:rollout_controls}

We first examine how the teacher-rollout advantage changes with initial-policy concentration under reverse KL and whether gains on target-like prefixes carry over to the prefixes visited by the final student (App.~\ref{app:rollout_transfer_boundary}). We then compare teacher utility on the ordinary and contradictory spatial tasks to connect the controlled-model results to the LLMs (App.~\ref{app:teacher_utility_bridge}).

\subsubsection{Initial concentration and deployment transfer}
\label{app:toy:rollout-sweep}
\label{app:rollout_transfer_boundary}

Unless stated otherwise, rollout-source comparisons use source-specific coupling rates selected by two-fold task-seed cross-fitting.

The main results show that teacher rollouts often improve acquisition, with the largest gains appearing in several contradictory LLM regimes. Because the main controlled comparison uses forward KL, we first test whether this qualitative preference persists under reverse KL. We repeat the $\lambda$ sweep on $96$ common task seeds. At $\kappa=0.6$, the teacher-minus-student acquisition difference rises from $+0.03$ points at $\lambda=0.625$ (95\% CI $[0.02,0.04]$) to $+1.21$ points at $\lambda=2.5$ ($[1.03,1.40]$), before falling to $+0.32$ points at $\lambda=5$ ($[-0.05,0.69]$). Thus concentration strengthens the teacher advantage over part of the range, but does not imply monotonic growth for every $\kappa$, consistent with the regime-dependent rollout-source behavior summarized in Sec.~\ref{sec:llm_results_rollout_src}. 
Moreover, in this $\kappa=0.6$ slice the selected coupling rates are identical for the teacher- and student-rollout endpoints in every held-out fold, so the contrast is unchanged by imposing a common rate. The observed preference is therefore not an artifact of source-specific coupling selection in this slice.

\begin{figure}[t]
    \centering
    \includegraphics[width=0.94\linewidth]{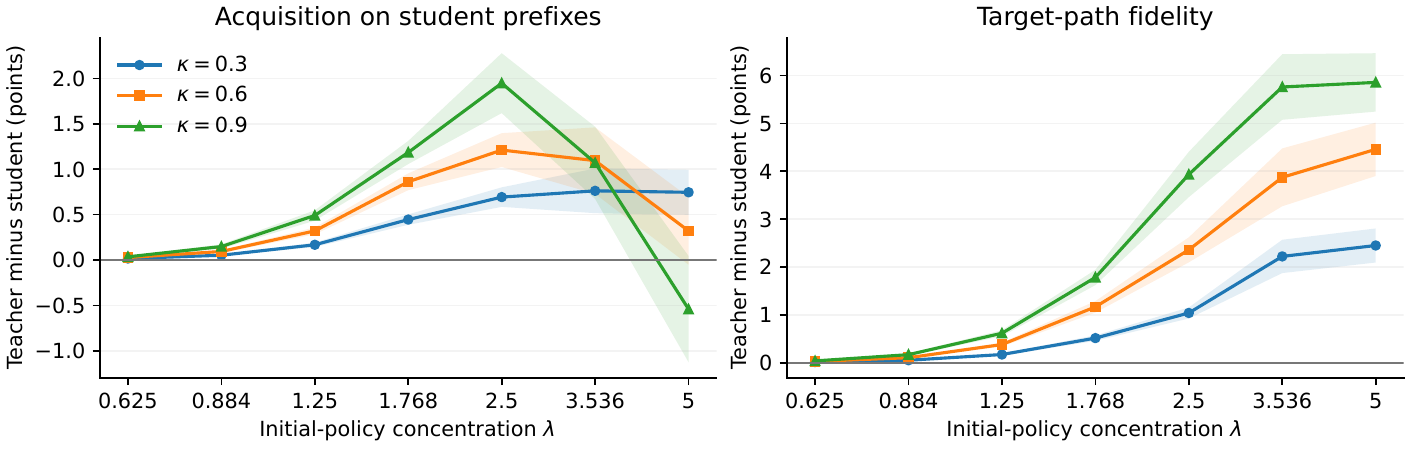}
    \caption{\textbf{Target-path fidelity and deployment-weighted acquisition can diverge.} Teacher-minus-student differences under reverse KL with $\rho_\phi=0.5$, $96$ paired task seeds, and coupling selected separately for each source by two-fold cross-fitting. Left: acquisition under each final student's own prefix occupancy. Right: target-path fidelity under the fixed target-policy occupancy. Bands show pointwise 95\% confidence intervals over paired seed-level contrasts.}
    \label{fig:rollout_transfer_boundary}
\end{figure}

Rollout source changes the prefix distribution on which distillation updates are taken. Training on teacher-generated trajectories can therefore improve target matching on teacher-emphasized prefixes without guaranteeing that this improvement transfers to the prefixes visited by the final student. Fig.~\ref{fig:rollout_transfer_boundary} separates these two effects.
We define \emph{target-path fidelity} as $\Delta_{\rm new}^{\rm tgt}$ from App.~\ref{app:toy:metrics}, which evaluates target matching under the fixed target-policy occupancy rather than the final student's occupancy. At $\lambda=5$ and $\kappa=0.9$, teacher rollouts improve target-path fidelity by $+5.86$ points (95\% CI $[5.25,6.47]$), while their deployment-weighted acquisition difference is $-0.54$ points ($[-1.13,0.04]$). Thus, teacher rollouts can produce substantially stronger target matching on target-like prefixes without yielding better behavior on the prefixes visited by the final student. This exposes the complementary advantage of student rollouts: by training directly on states encountered at deployment, they reduce the need for improvement learned under one trajectory distribution to transfer to another.

Teacher rollouts have higher mean deployment-weighted acquisition in $20/21$ cells of this sweep, reproducing the qualitative teacher-rollout preference summarized in Sec.~\ref{sec:llm_results_rollout_src}. At the same time, the deployment-alignment advantage identified in Fig.~\ref{fig:rollout_transfer_boundary} explains why this ordering need not be universal. However, none of the tested controlled-model settings produces a teacher-rollout advantage as large as those observed on the contradictory LLM tasks. This residual gap suggests that their rollout-source dependence also involves mechanisms absent from the controlled construction---for example, trajectory-dependent changes in how privileged context is used or representation-level effects---rather than prefix occupancy alone.

\subsubsection{Teacher utility on the student's path: from \texorpdfstring{$\lambda$}{lambda} to the contradictory tasks}
\label{app:teacher_utility_bridge}

Sec.~\ref{sec:llm_results_rollout_src} models a contradictory task by a larger initial-policy concentration $\lambda$ in the controlled model. We test this correspondence on the spatial pair (App.~\ref{app:llm:spatial-construction}), whose two tasks share all inputs $x$ and differ only in their reference responses and answers. Each reference response solves the problem as a sequence of moves, each written as a displacement such as $(+1, 0)$ (App.~\ref{app:llm:example-spatial}). For the $k$-th move, let $\mathbf v_k$ be its displacement under the ordinary convention, and $\mathbf v_k^\star$ the displacement that is correct for the task: $\mathbf v_k$ on spatial, $\mathbf v_k^{\rm rot}=R\mathbf v_k$ on spatial-contradiction (Eq.~\ref{eq:app:spatial-rotation}). We follow the reference response of the ordinary task, $y^{\mathrm{ref}}_{\rm ord}$, written in the convention the initial student follows, and let $t_k$ be the position where its $k$-th move starts. At each move, we record the probability that the initial teacher ($\theta_0^\star=\theta_0$, with context $c$) gives to the correct displacement, i.e., the product of the probabilities of its tokens:
\[
\widehat U_{\rm ctx}=\frac{1}{|\mathcal K|}\sum_{k\in\mathcal K}
\pi_{\theta_0}\!\left(\mathbf v_k^\star \mid x,c,\,y^{\mathrm{ref}}_{{\rm ord},<t_k}\right),
\]
where $\mathcal K$ contains the moves of the same $64$ training problems in both tasks, from the second move on, so that each scored move follows a move written in the ordinary convention. The student value is the same quantity without $c$. A value of $1$ would mean that the teacher always predicts the correct move from its context, regardless of the moves already written on the student’s path.

In the controlled model, the corresponding quantity is the teacher utility $U_{\rm ctx}(0)$ of App.~\ref{app:toy:metrics}: the similarity $\operatorname{Sim}(q_0,r_{\rm new})$, averaged over the prefixes the initial student visits. Two substitutions lead from it to $\widehat U_{\rm ctx}$. A move has one correct displacement, so the similarity to the target reduces to the probability of that displacement. The initial student writes its moves in the ordinary convention, so its prefixes are replaced by $y^{\mathrm{ref}}_{\rm ord}$: at the first move, it gives $0.19$ to the ordinary displacement and $0.01$ to the rotated one on Qwen ($0.12$ and $0.01$ on Ministral). This utility falls with $\lambda$. The initial teacher mixes the initial policy with the target, $q_0=(1-\kappa)p_0+\kappa r_{\rm new}$ (Sec.~\ref{sec:toy_model}), so raising $\lambda$ moves $p_0$ away from $r_{\rm new}$, and the teacher with it (Tab.~\ref{tab:teacher_utility_bridge}a). At the same time, the advantage of teacher rollouts grows (Fig.~\ref{fig:llm:rollout}b).

On the LLMs, the teacher utility on the student's path falls from $0.48$ on spatial to $0.15$ on spatial-contradiction for Qwen, and from $0.40$ to $0.16$ for Ministral, with disjoint 95\% intervals (Tab.~\ref{tab:teacher_utility_bridge}b). On its own path, the same teacher gives the correct move a probability of $0.39$ (Qwen) and $0.31$ (Ministral): the drop comes from the moves already written in the ordinary convention. On the same pair, the advantage of teacher rollouts is larger on the contradictory task (last column of Tab.~\ref{tab:teacher_utility_bridge}b). With respect to its ordinary twin, the contradictory task thus behaves like a larger $\lambda$ in the controlled model.

\begin{table}[h]
\centering
\small
\caption{\textbf{Teacher utility on the student's path falls with $\lambda$ in the controlled model and with contradiction in the LLMs.} (a)~$U_{\rm ctx}(0)$ at $\kappa=0.3$, as in Fig.~\ref{fig:llm:rollout}b.   (b)~Teacher: $\widehat U_{\rm ctx}$; student: the same quantity without $c$; brackets: 95\% bootstrap intervals over the $64$ problems. Last column: advantage of teacher rollouts, the median acquisition difference between teacher and student rollouts (Tab.~\ref{tab:app:stats-rollout}).}
\label{tab:teacher_utility_bridge}
\begin{tabular}{lcccc}
\multicolumn{5}{l}{(a) Controlled model}\\
\toprule
$\lambda$ & $0.625$ & $1.25$ & $2.5$ & $5$ \\
\midrule
$U_{\rm ctx}(0)$ & $0.82$ & $0.70$ & $0.57$ & $0.47$ \\
\bottomrule
\end{tabular}

\vspace{0.6em}

\begin{tabular}{llccc}
\multicolumn{5}{l}{(b) LLMs, spatial pair}\\
\toprule
Model & Task & Student & Teacher & Advantage of teacher rollouts (pp) \\
\midrule
Qwen      & Spatial               & $0.18$ & $0.48\ [0.39, 0.57]$ & $+2.4$ \\
          & Spatial-contradiction & $0.05$ & $0.15\ [0.10, 0.22]$ & $+4.9$ \\
\midrule
Ministral & Spatial               & $0.12$ & $0.40\ [0.33, 0.48]$ & $+2.5$ \\
          & Spatial-contradiction & $0.03$ & $0.16\ [0.12, 0.19]$ & $+4.4$ \\
\bottomrule
\end{tabular}
\end{table}

\subsection{Teacher coupling: useful supervision over a finite training budget}
\label{app:coupling_controls}

The coupling controls ask why the acquisition optimum occurs at an intermediate teacher-update rate, whether it persists under a complementary KL/rollout configuration, and which experimental conditions move that optimum. App.~\ref{coupling_failure} measures acquisition, retention cost, and contextual-teacher utility along the coupling sweep; App.~\ref{app:coupling_kl_rollout} changes both KL direction and rollout source; 
App.~\ref{app:context_corruption} varies contextual quality and training duration.

\subsubsection{Acquisition, retention cost, and teacher utility}
\label{coupling_failure}

\begin{figure}[t]
    \centering
    \includegraphics[width=0.98\linewidth]{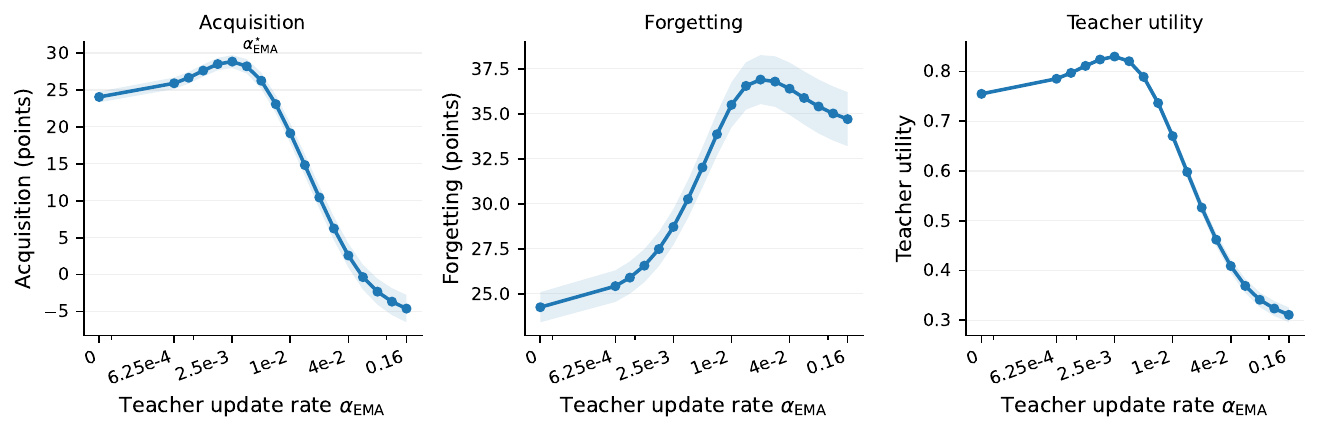}
    \caption{\textbf{Acquisition and contextual-teacher utility peak before fast coupling degrades adaptation.} Forward KL with teacher rollouts at $\lambda=2.5$, $\kappa=0.6$, and $\rho_\phi=0.5$, on $96$ task seeds. Panels report final acquisition, old-task retention loss $F_{\rm old}$, and teacher utility for separately trained coupling rates. Shading gives pointwise 95\% intervals over task seeds; the marked rate maximizes sample-mean acquisition on the tested grid.}
    \label{fig:coupling_forward_teacher_full}
\end{figure}

Fig.~\ref{fig:coupling_forward_teacher_full} complements the controlled panel of Fig.~\ref{fig:llm:coupling}b with the old-task retention cost $F_{\rm old}=-\Delta_{\rm old}$ and the contextual-teacher utility $U_{\rm ctx}$ defined in App.~\ref{app:toy:metrics}. The dedicated sweep uses forward KL, teacher rollouts, $\lambda=2.5$, $\kappa=0.6$, $\rho_\phi=0.5$, and $96$ task seeds. Moving from a frozen teacher to the sample-mean acquisition optimum $\alpha=0.0025$ raises acquisition from $24.05$ to $28.83$ points and teacher utility from $0.755$ to $0.830$, while retention loss increases from $24.25$ to $28.71$ points. At $\alpha=0.02$, acquisition falls to $10.45$ points and utility falls to $0.527$, while retention loss has risen to $36.89$ points; at $\alpha=0.16$, acquisition is $-4.59$ points. Moderate coupling can therefore improve the supervision available on the adapting student's final prefixes, but faster coupling eventually degrades acquisition while the retention cost remains high. Because $U_{\rm ctx}$ is weighted by each run's final student occupancy, it measures teacher utility on the states reached by that run rather than teacher quality on a common fixed prefix distribution.

\subsubsection{Persistence under reverse KL and student rollouts}
\label{app:coupling_kl_rollout}

\begin{figure}[t]
    \centering
    \includegraphics[width=0.72\linewidth]{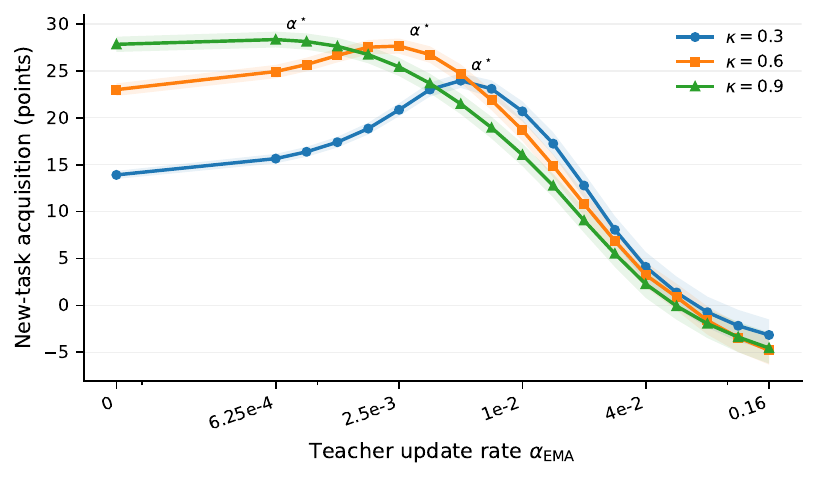}
    \caption{\textbf{The coupling peak persists with reverse KL and student rollouts.} Acquisition at $\lambda=2.5$ and $\rho_\phi=0.5$ on $96$ task seeds. Increasing $\kappa$ shifts the cross-fitted selected rate toward slower teacher updates. Shading gives pointwise 95\% intervals at each fixed rate.}
    \label{fig:coupling_reverse_student}
\end{figure}

Fig.~\ref{fig:coupling_reverse_student} repeats the coupling sweep under reverse KL with student rollouts, using the same $96$-seed cohort, $\lambda=2.5$, and $\rho_\phi=0.5$. Acquisition again rises and then falls with $\alpha$. For $\kappa=0.3,0.6,0.9$, task-seed cross-fitting selects $\alpha^\star=0.005$, $0.0025$, and $0.000625$, respectively, with mean held-out acquisitions of $23.96$, $27.65$, and $28.34$ points. The finite useful-coupling window and its shift toward slower updates as contextual correction strengthens therefore persist under this complementary joint change of KL direction and rollout source. Because both method axes change together, this control establishes persistence of the coupling pattern rather than isolating the contribution of either axis individually.

\subsubsection{What moves the useful coupling timescale?}
\label{app:context_corruption}
\label{app:coupling_horizon}

Two stress tests show that the numerical optimum is not an intrinsic constant of the EMA update. First, we corrupt the privileged feature displacement while keeping the reverse-KL/student-rollout setting of Fig.~\ref{fig:coupling_reverse_student}. For each task seed, collect the clean displacements into the state-by-feature matrix $\Psi_\kappa$ and draw a Gaussian matrix $G$ of the same shape. We construct a fixed perturbation orthogonal to $\Psi_\kappa$ in the Frobenius inner product:
\[
G_\perp=G-\frac{\langle G,\Psi_\kappa\rangle_F}{\|\Psi_\kappa\|_F^2}\Psi_\kappa,\qquad
\widetilde{\Psi}_\kappa=\Psi_\kappa+\sigma_\phi\frac{\|\Psi_\kappa\|_F}{\|G_\perp\|_F}G_\perp.
\]
The perturbed contextual displacement at each prefix is the corresponding row of $\widetilde{\Psi}_\kappa$. Orthogonality and relative noise magnitude are imposed jointly over all state-feature entries within each seed, and the perturbation remains fixed during training. The $64$-seed control selects the sample-mean acquisition optimum over the same 18-rate coupling grid. Tab.~\ref{tab:toy:coupling-stress-tests}a shows that relative noise $0.5$ leaves the clean ordering unchanged; at scale $1$, the preferred rate slows for all three $\kappa$ values and freezing is already optimal at $\kappa=0.9$; at scale $2$, freezing maximizes acquisition for all three. Thus $\kappa$ calibrates the clean initial correction, but does not determine the useful coupling rate once the privileged signal itself is corrupted.

Second, we vary the training budget on $48$ task seeds at $\lambda=2.5$, $\kappa=0.6$, and $\rho_\phi=0.5$. All four KL-direction/rollout-source combinations select the same sample-mean optimum over $\alpha\in\{0,0.00125,0.0025,0.005,0.01\}$: $0.005$, $0.0025$, and $0.00125$ at $100$, $200$, and $400$ updates, respectively, giving $\alpha^\star N=0.5$ in each case. At $800$ updates, $0.00125$ is again selected, but this is the smallest positive rate in the tested duration grid; the experiment therefore does not resolve whether a slower rate would be better. The approximate inverse scaling over $100$--$400$ updates should not be read as a general law.

\begin{table}[t]
\centering
\small
\caption{\textbf{Context quality and training duration move the useful coupling rate.} Entries are sample-mean acquisition-maximizing tested rates. (a) Feature-space corruption uses reverse KL, student rollouts, $\lambda=2.5$, $\rho_\phi=0.5$, and $64$ task seeds. (b) Duration uses $\lambda=2.5$, $\kappa=0.6$, $\rho_\phi=0.5$, and $48$ seeds; all four KL/rollout combinations select the same rate.}
\label{tab:toy:coupling-stress-tests}
\begin{tabular}{@{}lccc@{}}
\toprule
\multicolumn{4}{c}{(a) Context corruption}\\
\cmidrule(lr){1-4}
Relative noise $\sigma_\phi$ & $\kappa=0.3$ & $\kappa=0.6$ & $\kappa=0.9$\\
\midrule
$0$   & $0.005$    & $0.0025$  & $0.000625$\\
$0.5$ & $0.005$    & $0.0025$  & $0.000625$\\
$1$   & $0.00354$  & $0.00125$ & $0$\\
$2$   & $0$        & $0$       & $0$\\
\midrule
\multicolumn{4}{c}{(b) Training duration}\\
\cmidrule(lr){1-4}
Updates $N$ & Selected $\alpha^\star$ & $\alpha^\star N$ & Note\\
\midrule
$100$ & $0.005$   & $0.5$ & \\
$200$ & $0.0025$  & $0.5$ & \\
$400$ & $0.00125$ & $0.5$ & \\
$800$ & $0.00125$ & $1.0$ & grid-censored \\%
\bottomrule
\end{tabular}
\end{table}

\subsection{KL direction: retention costs and regime reversals}
\label{app:kl_controls}

The KL analysis first documents the exact endpoints used in the main controlled comparison (App.~\ref{cross_axis_tradeoffs}), then separates input-side feature sharing from output-side readout compatibility (App.~\ref{app:rho_phi}), maps the region in which forward KL improves both acquisition and retention (App.~\ref{app:kl_regimes}), and finally tests teacher entropy as a complementary regime diagnostic (App.~\ref{app:kl_entropy}).

\subsubsection{Endpoints behind the main KL comparison}
\label{cross_axis_tradeoffs}

Fig.~\ref{fig:llm:axis-tradeoffs}b summarizes a change in the KL-direction trade-off between the reference regime and a regime with stronger contextual correction and complete feature sharing. Here we unpack the frozen and coupling-selected endpoints underlying those two comparisons. This separates the gain provided by teacher coupling from the remaining forward--reverse difference, and makes clear how the retention advantage of reverse KL at the reference setting gives way to a joint forward-KL advantage in the high-$\kappa$, high-$\rho_\phi$ regime. Tab.~\ref{tab:toy:kl-endpoints} gives both the frozen and selected endpoints. At $(\kappa,\rho_\phi)=(0.6,0.5)$, tuning reverse-KL coupling raises acquisition from $24.00$ to $29.43$ points while retention changes from $-20.55$ to $-24.81$ points, i.e. $+5.43$ acquisition for $4.26$ points of additional retention loss. The selected forward endpoint acquires $0.26$ points less and retains $3.04$ points less than the selected reverse endpoint. At $(0.9,1)$, the selected forward endpoint instead exceeds reverse by $1.23$ acquisition and $0.38$ retention points in mean. 

\begin{table}[t]
\centering
\small
\caption{\textbf{Controlled-model endpoints behind Fig.~\ref{fig:llm:axis-tradeoffs}b.} Acquisition and retention are reported in percentage points, so higher is better on both columns. Rates are selected separately for each KL direction by sample-mean acquisition over the $96$ common task seeds.}
\label{tab:toy:kl-endpoints}
\begin{tabular}{@{}cccrr@{}}
\toprule
$(\kappa,\rho_\phi)$ & KL & Endpoint / $\alpha$ & Acquisition & Retention\\
\midrule
$(0.6,0.5)$ & Reverse & frozen / $0$ & $24.00$ & $-20.55$\\
             & Reverse & selected / $0.0025$ & $29.43$ & $-24.81$\\
             & Forward & frozen / $0$ & $24.43$ & $-23.63$\\
             & Forward & selected / $0.0025$ & $29.17$ & $-27.84$\\
\midrule
$(0.9,1)$   & Reverse & frozen / $0$ & $29.80$ & $-39.17$\\
             & Reverse & selected / $0.000625$ & $30.61$ & $-41.16$\\
             & Forward & frozen / $0$ & $30.48$ & $-38.98$\\
             & Forward & selected / $0.000884$ & $31.84$ & $-40.78$\\
\bottomrule
\end{tabular}
\end{table}

\subsubsection{Feature sharing and readout compatibility}
\label{app:rho_phi}

We next separate two notions of task relatedness in the construction. Feature overlap $\rho_\phi$ controls how strongly the old and new tasks reuse the same input features, whereas readout compatibility $\rho_W$ controls the alignment of the auxiliary target-generating readout with the pretrained readout (App.~\ref{app:toy:task-geometry}). Panel~(a) of Tab.~\ref{tab:toy:feature-readout-controls} uses the same teacher-rollout, $96$-seed main grid as Fig.~\ref{fig:llm:axis-tradeoffs}b, with $\lambda=2.5$, $\kappa=0.6$, and fixed $\alpha=0.0025$. Raising $\rho_\phi$ from $0$ to $1$ increases the retention cost under both KL directions and compresses the reverse-KL retention advantage from $3.19$ points to essentially zero. The paired forward-minus-reverse retention interval is $[-3.83,-2.55]$ points at $\rho_\phi=0$, $[-3.76,-2.32]$ at $0.5$, and $[-0.27,0.26]$ at $1$. Complete feature sharing therefore erodes the reference retention separation; the near-zero endpoint is not evidence of statistical equivalence or of a forward-KL retention advantage by itself.

\paragraph{Readout-compatibility control.}\label{rho_W_ablation}
Panel~(b) tests whether the KL comparison depends on the auxiliary target-readout compatibility $\rho_W$, using an independent $64$-seed cohort at the reference task geometry. With a frozen teacher, forward KL acquires more across all seven tested $\rho_W$ values, while reverse KL retains more throughout. At the moderate coupling rate $\alpha=0.0025$, varying $\rho_W$ can reverse the small acquisition difference between the two KL directions, but the reverse-KL retention advantage again persists across the entire sweep. Thus readout compatibility can modulate which KL direction acquires slightly more, particularly once the teacher is coupled, without changing the more stable retention ordering emphasized in the main results.

\begin{table}[t]
\centering
\small
\caption{\textbf{Feature sharing and output-readout compatibility are distinct controls.} ``Rev. ret. adv.'' is reverse minus forward retention, in points. (a) Main-grid teacher-rollout slice with $96$ seeds, $\lambda=2.5$, $\kappa=0.6$, and fixed $\alpha=0.0025$. (b) Readout-compatibility control with $64$ seeds under the reference task geometry; each entry reports $\alpha=0.0025$ / frozen ($\alpha=0$). Panel~(b) reports mean contrasts without confidence intervals.}
\label{tab:toy:feature-readout-controls}
\begin{tabular}{@{}crrrr@{}}
\toprule
\multicolumn{5}{c}{(a) Feature overlap}\\
\cmidrule(lr){1-5}
$\rho_\phi$ & Fwd$-$Rev acq. & Fwd retention & Rev retention & Rev. ret. adv.\\
\midrule
$0$   & $-0.18$ & $-22.91$ & $-19.72$ & $3.19$\\
$0.5$ & $-0.26$ & $-27.84$ & $-24.81$ & $3.04$\\
$1$   & $+0.02$ & $-40.00$ & $-40.00$ & $0.00$\\
\bottomrule
\end{tabular}

\vspace{1mm}
\begin{tabular}{@{}crr@{}}
\toprule
\multicolumn{3}{c}{(b) Readout compatibility: $\alpha=0.0025$ / frozen}\\
\cmidrule(lr){1-3}
$\rho_W$ & Fwd$-$Rev acq. & Rev. ret. adv.\\
\midrule
$-0.9$   & $+0.21\,/\,+0.77$ & $3.98\,/\,4.13$\\
$-0.707$ & $+0.10\,/\,+0.72$ & $3.66\,/\,3.97$\\
$-0.5$   & $-0.02\,/\,+0.66$ & $3.49\,/\,3.78$\\
$0$      & $-0.23\,/\,+0.59$ & $3.28\,/\,3.42$\\
$0.5$    & $-0.23\,/\,+0.56$ & $3.04\,/\,3.14$\\
$0.707$  & $-0.23\,/\,+0.55$ & $2.86\,/\,2.93$\\
$0.9$    & $-0.18\,/\,+0.53$ & $2.78\,/\,2.85$\\
\bottomrule
\end{tabular}

\end{table}

\subsubsection{Where the KL ordering reverses}
\label{app:kl_regimes}

To map the transition more densely than the $96$-seed main grid, we use a dedicated teacher-rollout interaction sweep with learning rate $10^{-3}$, $200$ updates, and $64$ common task seeds. We cross
\[
\begin{aligned}
\rho_\phi&\in\{0,0.259,0.5,0.707,0.866,0.966,1\},\\
\lambda&\in\{0.625,0.884,1.25,1.768,2.5,3.536,5\},\\
\kappa&\in\{0.3,0.6,0.9\},
\end{aligned}
\]
Displayed values of $\rho_\phi$ and $\lambda$ are rounded. With the restricted coupling grid $\alpha\in\{0,0.00125,0.0025,0.005\}$, the sweep contains $588$ paired forward/reverse configurations. This restriction avoids the fast-coupling tail illustrated in App.~\ref{coupling_failure}.

For configuration $\theta=(\rho_\phi,\lambda,\kappa,\alpha)$ and task seed $j$, define the forward-minus-reverse acquisition and retention contrasts
\[
\delta A_j(\theta)=100\!\left[\Delta_{{\rm new},F}^{(j)}(\theta)-\Delta_{{\rm new},R}^{(j)}(\theta)\right],\qquad
\delta R_j(\theta)=100\!\left[\Delta_{{\rm old},F}^{(j)}(\theta)-\Delta_{{\rm old},R}^{(j)}(\theta)\right],
\]
and let $\overline{\delta A}(\theta)$ and $\overline{\delta R}(\theta)$ denote their seed means. A \emph{retention inversion} has $\overline{\delta R}>0$; a \emph{joint forward advantage} additionally requires $\overline{\delta A}>0$. Connectivity is defined on the sampled four-dimensional grid, with nearest neighbors differing by one consecutive sampled value of exactly one coordinate.

Tab.~\ref{tab:toy:kl-regime}a summarizes the topology. All $185$ retention-inversion cells form one connected component. Joint forward advantage occurs in $46$ cells, $45$ of which form one component. The isolated mean-sign cell is $(\rho_\phi,\lambda,\kappa,\alpha)\simeq(0.259,1.25,0.9,0.005)$; its acquisition contrast is only $+0.060$ points (95\% CI $[-0.073,0.192]$) and its retention contrast $+0.123$ points ($[-0.128,0.374]$). Requiring the pointwise 95\% intervals of both contrasts to lie above zero leaves $31$ cells, all in one component. These intervals are pointwise and do not provide simultaneous confidence for the entire region.

The joint region is concentrated at stronger contextual correction and higher feature sharing: no joint cell occurs at $\kappa=0.3$, compared with $8$ at $0.6$ and $38$ at $0.9$. Its size also increases across the restricted coupling grid, from $3$ cells at $\alpha=0$ to $9$, $13$, and $21$ at $0.00125$, $0.0025$, and $0.005$. Membership in the $31$-cell interval-supported subset is monotone along the sampled $\rho_\phi$, $\kappa$, and $\alpha$ coordinates, whereas no general monotone ordering holds in $\lambda$. Tab.~\ref{tab:toy:kl-regime}b shows the sampled overlap boundary at $\kappa=0.9$: stronger coupling generally lowers the overlap at which joint improvement first appears. 

\begin{table}[t]
\centering
\small
\caption{\textbf{Topology and sampled boundary of the forward-favorable KL regime.} Teacher rollouts, $64$ common seeds, and the $588$-cell restricted interaction grid defined in the text. Panel (a) distinguishes retention-only inversion from joint forward improvement. Panel (b) gives the smallest tested $\rho_\phi$ with joint mean improvement at $\kappa=0.9$ for the four largest displayed concentrations; a dash means that no tested overlap qualifies.}
\label{tab:toy:kl-regime}
\begin{tabular}{@{}lrr@{}}
\toprule
\multicolumn{3}{c}{(a) Region topology}\\
\cmidrule(lr){1-3}
Criterion & Cells & Component sizes\\
\midrule
Forward retains more & $185$ & $185$\\
Forward acquires and retains more & $46$ & $45+1$\\
Both pointwise 95\% CIs $>0$ & $31$ & $31$\\
\midrule
\multicolumn{3}{c}{(b) Smallest tested $\rho_\phi$ with joint mean improvement, $\kappa=0.9$}\\
\bottomrule
\end{tabular}

\vspace{1mm}
\begin{tabular}{@{}ccccc@{}}
\toprule
$\alpha$ & $\lambda\simeq1.768$ & $\lambda=2.5$ & $\lambda\simeq3.536$ & $\lambda=5$\\
\midrule
$0$       & --      & $1$     & $1$     & $1$\\
$0.00125$ & $0.866$ & $0.966$ & $0.966$ & $0.966$\\
$0.0025$  & $0.866$ & $0.866$ & $0.966$ & $0.966$\\
$0.005$   & $0.707$ & $0.866$ & $0.866$ & $0.966$\\
\bottomrule
\end{tabular}
\end{table}

\subsubsection{Teacher uncertainty as a complementary regime diagnostic}
\label{app:kl_entropy}

On the same $64$-seed interaction sweep, define the initial contextual-teacher entropy under the initial student's prefix occupancy by
\[
H_0=\frac1T\sum_{s\in\mathcal S}\nu_{p_0}(s)
\left[-\sum_a q_0(a\mid s)\log q_0(a\mid s)\right].
\]
We average $H_0$ over task seeds and $\overline{\delta R}$ over the four restricted coupling rates, giving one observation for each of the $147$ $(\rho_\phi,\lambda,\kappa)$ geometries. Across these observations, Spearman's $r_s=0.700$: higher-entropy geometries tend to shift the retention ordering toward forward KL. The association remains positive when overlap is fixed ($r_s=0.649$, $0.734$, and $0.792$ for $\rho_\phi=0$, $0.5$, and $1$, respectively), so it is not solely produced by pooling overlap levels.

Entropy nevertheless does not determine the ordering by itself. At $\lambda=2.5$ and $\rho_\phi=0.5$, increasing $\kappa$ from $0.6$ to $0.9$ raises $H_0$ from $1.871$ to $2.028$ nats, yet reverse KL still retains more at fixed $\alpha=0.0025$. Holding $\lambda=2.5$ and $\kappa=0.9$ instead and increasing $\rho_\phi$ from $0.5$ to $1$ changes $H_0$ by less than $0.001$ nats, while the forward-minus-reverse retention contrast changes from $-2.37$ to $+2.54$ points. These contrasts are consistent with teacher uncertainty marking the supervision regime and feature sharing influencing its retention consequences, but they do not identify reverse-KL mode seeking as the cause of the inversion.

For the LLM diagnostic, under frozen-teacher rollouts at learning rate $10^{-5}$, mean teacher-token entropy over the first ten logged steps is $2.08$ nats for Ministral and $0.31$ for Qwen across the five tasks, a paired task-level difference of $1.77$ nats (95\% CI $[1.11,2.44]$). Because the models use different tokenizers and generate different prefixes, these absolute values are descriptive rather than directly calibrated across models. Consistent with this interpretation, \citet{jin2026entropyopd} report reduced generation diversity and unstable reverse-KL learning signals under high-entropy teachers. Their result concerns diversity and optimization rather than old-task retention, but is consistent with teacher uncertainty marking a distinct KL regime. Our controls show, however, that uncertainty alone does not determine the retention ordering. 

\subsection{Construction and numerical checks}
\label{app:robustness}

\begin{table}[t]
\centering
\scriptsize
\setlength{\tabcolsep}{2.6pt}
\caption{\textbf{Scoped numerical and structural checks.} Each sweep uses $48$ task seeds, with $\lambda=2.5$, $\kappa=0.6$, $\rho_\phi=0.5$, learning rate $10^{-3}$, and $200$ updates. Structural sweeps vary one coordinate of the reference $(T,K,D)=(4,8,64)$ at a time and use separate seed cohorts, so their reference rows need not coincide. The four contrasts are defined in the text; all values are points. Dashes denote contrasts not reported for the estimator sweep.}
\label{tab:toy:robustness}
\begin{tabular}{@{}llrrrr@{}}
\toprule
Check & Value & Rev. ret. & Teac.-roll & Coupling & Fwd$-$Rev\\
      &       & adv. & adv. & gain & acq.\\
\midrule
\multicolumn{6}{c}{(a) Prefix-occupancy estimator}\\
\cmidrule(lr){1-6}
Estimator & exact   & $2.75$ & -- & $4.80$ & --\\
          & MC-64   & $2.12$ & -- & $4.58$ & --\\
          & MC-256  & $3.35$ & -- & $4.70$ & --\\
          & MC-1024 & $3.30$ & -- & $4.85$ & --\\
\midrule
\multicolumn{6}{c}{(b) Structural variants}\\
\cmidrule(lr){1-6}
$D$ & $32$  & $2.02$ & $0.85$ & $3.60$ & $-0.50$\\
    & $64$  & $2.89$ & $1.17$ & $4.98$ & $+0.35$\\
    & $128$ & $3.62$ & $1.52$ & $5.85$ & $+0.66$\\
$T$ & $3$   & $3.74$ & $1.53$ & $7.09$ & $+0.06$\\
    & $4$   & $3.23$ & $1.22$ & $4.80$ & $+0.49$\\
    & $5$   & $3.80$ & $1.27$ & $3.97$ & $+0.38$\\
$K$ & $6$   & $3.75$ & $1.61$ & $4.94$ & $+0.39$\\
    & $8$   & $3.33$ & $1.45$ & $4.55$ & $+0.03$\\
    & $12$  & $3.63$ & $1.36$ & $4.48$ & $+0.63$\\
\bottomrule
\end{tabular}
\end{table}

Finally, we test two conspicuous simplifications of the controlled construction: exact prefix occupancy and the finite choices of response horizon, vocabulary size, and feature dimension. These checks delimit the scope of the mechanism results rather than establishing parameter-invariant behavior. Tab.~\ref{tab:toy:robustness} uses four contrasts at the reference task geometry. The \emph{reverse retention advantage} is reverse-minus-forward retention under student rollouts at fixed $\alpha=0.0025$; the \emph{teacher rollout advantage} is teacher-minus-student acquisition under reverse KL at the same rate; the \emph{coupling gain} is reverse-KL/student-rollout acquisition at $\alpha=0.0025$ minus its frozen-teacher value; and the final column is forward-minus-reverse acquisition under student rollouts at $\alpha=0.0025$.

Replacing exact occupancy by Monte Carlo estimates with $64$, $256$, or $1024$ sampled trajectories preserves both the moderate-coupling gain and the sign of the reverse retention advantage. In separate one-factor sweeps over $T\in\{3,4,5\}$, $K\in\{6,8,12\}$, and $D\in\{32,64,128\}$, the mean reverse retention advantage, teacher-rollout acquisition advantage, and moderate-coupling acquisition gain remain positive. The absolute forward-minus-reverse acquisition contrast is less stable: changing $D$ moves it from $-0.50$ to $+0.66$ points. Taken together, these checks show that the rollout-source, coupling, and reverse-KL retention effects are robust to the tested estimator and structural choices, whereas the acquisition ordering between KL directions remains regime-dependent rather than a fixed property of the divergence.